\documentclass[10pt, journal, compsoc]{IEEEtran}
\usepackage{bm}
\usepackage{amsmath}
\usepackage{amssymb}
\usepackage{subfigure}
\usepackage{diagbox}
\usepackage{adjustbox}
\usepackage{makecell}
\usepackage{mathrsfs}
\usepackage{xcolor}
\usepackage{balance}
\usepackage[citecolor=blue, colorlinks]{hyperref}
\usepackage[T1]{fontenc}
\usepackage{booktabs}

\newcommand{\mr}[1]{\textcolor{black}{#1}} 

\begin{document}

\title{Flattening-Net: Deep Regular 2D Representation for 3D Point Cloud Analysis}

\author{
     Qijian Zhang, Junhui Hou, Yue Qian, Yiming Zeng, Juyong Zhang, and Ying He 
\IEEEcompsocitemizethanks{\IEEEcompsocthanksitem Q. Zhang, J. Hou, Y. Qian, and Y. Zeng are with the Department of Computer Science, City University of Hong Kong, Hong Kong SAR. Email: qijizhang3-c@my.cityu.edu.hk; jh.hou@cityu.edu.hk; yueqian4-c@my.cityu.edu.hk; ym.zeng@my.cityu.edu.hk;
\IEEEcompsocthanksitem J. Zhang is with the School of Mathematical Sciences,
University of Science and Technology of China, Hefei, Anhui, 230026 China. Email: juyong@ustc.edu.cn;
\IEEEcompsocthanksitem Y. He is with the School of Computer Science and Engineering, Nanyang
Technological University, Singapore, 639798. Email: yhe@ntu.edu.sg;
}
\thanks{This work was supported in part by the Hong Kong Research Grants Council under Grant  11202320, Grant 11219422, and Grant 11218121, and in part by the Natural Science Foundation of China under Grant  61871342. Corresponding author: Junhui Hou 
}}


\IEEEtitleabstractindextext{
\begin{abstract} Point clouds are characterized by irregularity and unstructuredness, which pose challenges in efficient data exploitation and discriminative feature extraction. In this paper, we present an unsupervised deep neural architecture called Flattening-Net to represent irregular 3D point clouds of arbitrary geometry and topology as a completely regular 2D point geometry image (PGI) structure, in which coordinates of spatial points are captured in colors of image pixels. \mr{Intuitively, Flattening-Net implicitly approximates a locally smooth 3D-to-2D surface flattening process  while effectively preserving neighborhood consistency.} \mr{As a generic representation modality, PGI inherently encodes the intrinsic property of the underlying manifold structure and facilitates surface-style point feature aggregation.} To demonstrate its potential, we construct a unified learning framework directly operating on PGIs to achieve \mr{diverse types of high-level and low-level} downstream applications driven by specific task networks, including classification, segmentation, reconstruction, and upsampling. Extensive experiments demonstrate that our methods perform favorably against the current state-of-the-art competitors. We will make the code and data  publicly available at \url{https://github.com/keeganhk/Flattening-Net}.
\end{abstract}

}

\maketitle

\IEEEraisesectionheading{\section{Introduction} \label{introduction}}
\IEEEPARstart{T}{he} recent decade has witnessed remarkable advancement of deep learning architectures, especially convolutional neural networks (CNNs), in analyzing regular visual modalities, such as 2D images/videos and 3D volumes \cite{krizhevsky2012imagenet, karpathy2014large, tran2015learning, long2015fully, he2016deep}, where powerful convolution operations can be naturally and efficiently performed on such dense and uniform grid structures. However, it is highly non-trivial to adapt the existing learning frameworks that have been well-developed in regular data domains to the unstructured point cloud modality that is typically characterized by sparsity, irregularity, and unorderedness. Different from regular-domain visual understanding scenarios where a variety of off-the-shelf backbone feature extraction networks \cite{simonyan2014very, szegedy2015going, szegedy2016rethinking, he2016deep, xie2017aggregated, huang2017densely, tan2019efficientnet} are available, such that researchers can devote themselves to designing downstream processing units for specific visual tasks, there is still a lack of generic and mature feature learning paradigms for point clouds.

One of the most straightforward strategies to deal with irregularity is to pre-transform raw point clouds to regular data representation structures through rasterization. \textit{Volumetric models} \cite{maturana2015voxnet, qi2016volumetric, wu20153d} are a representative family of such processing pipelines, in which spatial points are quantized into occupancy grids, such that standard 3D convolutions can be directly applied for feature extraction. However, due to the sparsity of point cloud data, most computational resources can be wasted on empty voxels, which becomes the major computational bottleneck. Moreover, limited by the cubic growth of computational complexity and memory footprint, the voxelization-based paradigm only applies to low-resolution volumes. The follow-up works further introduce octree \cite{riegler2017octnet, wang2017cnn} and kd-tree \cite{klokov2017escape} based hierarchical adaptive indexing structures to reduce computational and memory costs, which become applicable to high-resolution volumes and allow to capture geometric details. Alternatively, \textit{projective models} \cite{su2015multi, qi2016volumetric, kalogerakis20173d, yu2018multi, kanezaki2018rotationnet} represent a 3D shape by multiple 2D view images, and then aggregate view-wise features extracted from mature 2D CNNs to form a global shape descriptor. Despite its dominating performance in classification/retrieval tasks, the projection-based paradigm relies on external large-scale image databases for pretraining and is not directly applicable to fine-grained 3D understanding tasks requiring point-wise prediction, such as semantic segmentation and normal estimation.

\begin{figure*}[t]
	\centering
	\includegraphics[width=0.975\linewidth]{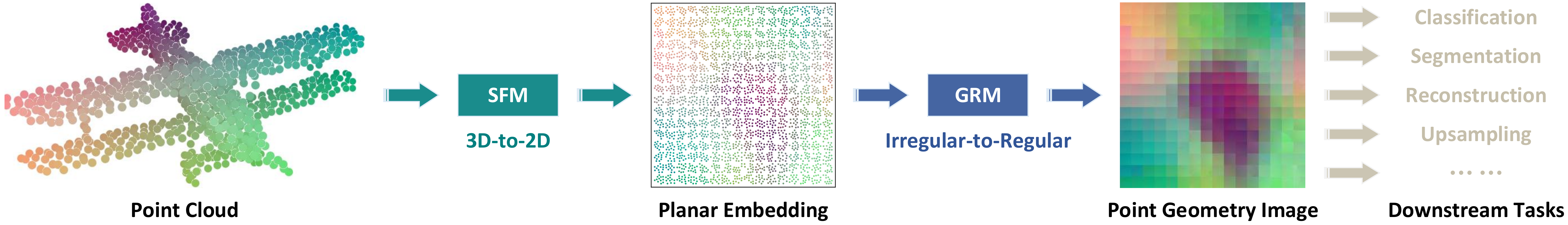}
	\caption{Given an irregular point cloud of arbitrary geometric and topological structures, Flattening-Net generates a regular point geometry image (PGI) that encodes point coordinates as pixel colors while preserving local neighborhood consistency effectively. In the first stage, the input 3D points are embedded onto a 2D unit square domain through the SFM module. In the second stage, we resample the embedding points on lattice grids to produce a regular PGI. As a generic representation modality for point clouds, PGI naturally fits into a rich variety of \mr{high-level and low-level downstream applications driven by specific task networks.}}
	\label{fig:overall-workflow}
\end{figure*}

Without adopting rasterization as a pre-processing procedure for structure conversion, \textit{point-based models} \cite{qi2017pointnet, qi2017pointnet++,groh2018flex,hua2018pointwise, li2018so, su2018splatnet, xu2018spidercnn, li2018pointcnn, wu2019pointconv, liu2019relation, thomas2019kpconv, zhang2019shellnet} directly take point sets as input and can produce point-wise embeddings conveniently. As a pioneering work, PointNet \cite{qi2017pointnet} is built upon shared multi-layer perceptrons (MLPs) that lift point coordinates to high-dimensional features, and then extracts a global codeword by applying channel-wise max-pooling. The follow-up works are devoted to tailoring convolution-like operators defined on point sets and mimicking conventional CNN architectures. Besides, since graph convolution is naturally suitable for irregular data modeling, there also exist a family of specialized \textit{graph-based models} \cite{verma2018feastnet, wang2019dynamic} that demonstrate great potential in point cloud feature learning. However, as pointed out in \cite{liu2019pvcnn, xu2020grid}, most computational resources in point-based models are wasted on the process of neighborhood construction (e.g., $k$-NN) and structurization, rather than on the actual feature extraction, which can significantly limit model efficiency as the number of input points increases. Moreover, point-based models typically suffer from unsatisfactory scalability and generality when dealing with varying number of input points, which means that one should carefully configure network structures and hyperparameters, (\textit{e.g.}, neighborhood size) to accommodate input data with different number of points.

\begin{figure*}[t]
	\centering
	\includegraphics[width=0.975\linewidth]{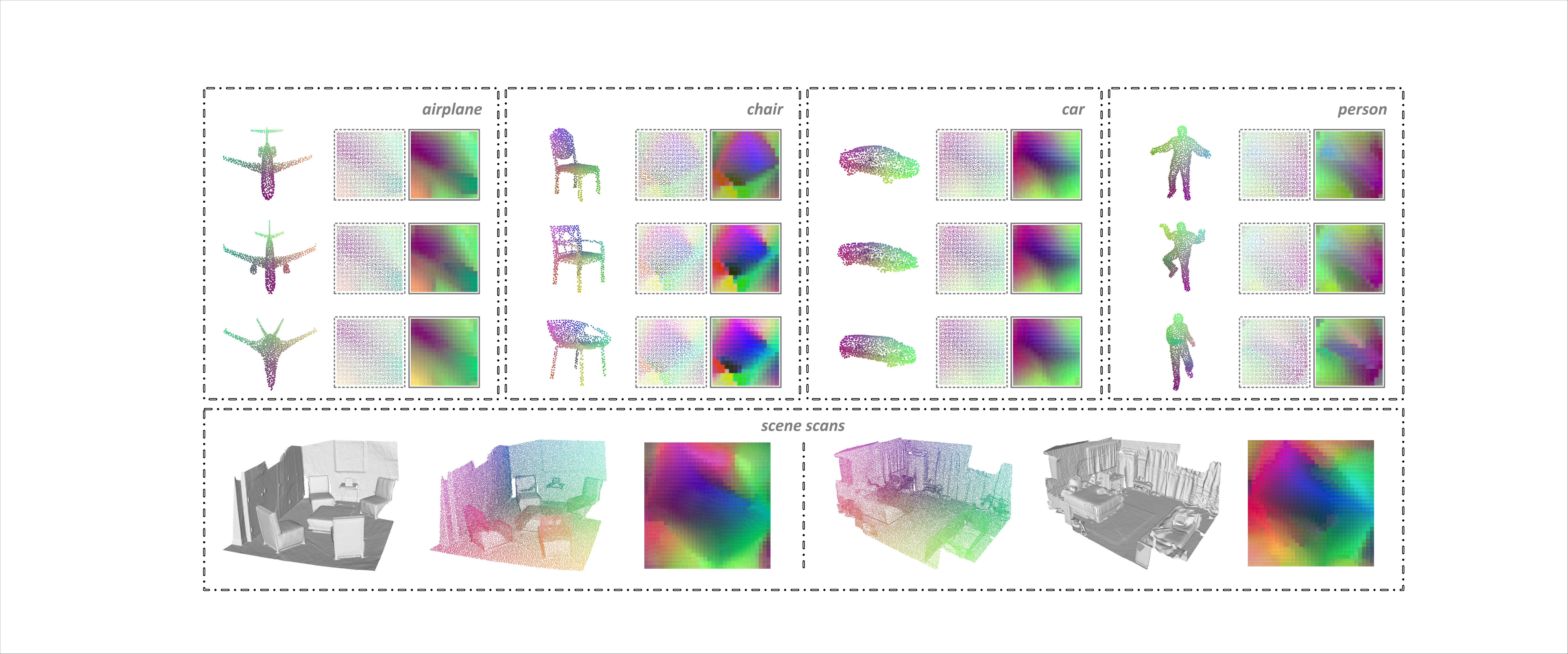}
	\caption{\mr{Visualization of 2D PGI representation structures generated from diverse object-level and scene-level 3D point clouds, where we colorize input points, planar embeddings, and the resulting PGIs according to their correspondence relationships, such that pixel colors uniquely reflect spatial positions. It is observed that the generated PGIs have smooth color distributions, indicating that spatial continuity is effectively maintained during such a surface-to-plane mapping process.}}
	\label{fig:pgi-gallery}
\end{figure*}

In this paper, we seek to represent an unstructured point cloud by a regular grid structure dubbed as point geometry image (PGI), which encodes Cartesian coordinates of points as color values of image pixels. Intuitively, we can interpret the structural conversion between point clouds and PGIs as \mr{a locally smooth deformation} between irregular points on the target 3D surface and regular grids on the pre-defined 2D lattice, which can meaningfully unfold and flatten 3D surfaces onto 2D planar domains. Technically, as illustrated in Figure~\ref{fig:overall-workflow}, we present an unsupervised deep neural architecture called Flattening-Net, a two-stage learning framework composed of a surface flattening module (SFM) and a grid resampling module (GRM). In the first stage, we explore two complementary learning paradigms for building \mr{locally smooth mappings} between 3D object surfaces and 2D planar domains, based on which we design the SFM for generating planar flattenings from the original point cloud. In the second stage, we redistribute the irregular embedding points to dense grid positions of a uniform lattice by solving an assignment problem, which can deduce a three-channel image structure. \mr{Figure \ref{fig:pgi-gallery} visualizes a gallery of PGIs generated from various 3D object models and scene scans.}

Driven by the structural characteristics of PGIs, we customize concentric-square convolution (CSConv), an efficient feature extraction operator that is suitable for aggregating statistics on local manifolds, to produce vectorized regional descriptors in a unified and scalable fashion. We also investigate specific task networks coupled with CSConv to achieve downstream applications, including classification, segmentation, reconstruction, and upsampling. It is worth noting that our major focus is to enrich the application scenarios and to reveal the universality of the PGI representation, rather than to customize highly-specialized processing techniques for pursuing the best performance in all evaluation tasks involved in experiments. Still, it turns out that our methods can perform favorably against the current state-of-the-art competitors.

In general, our processing pipeline can be divided into two separate stages: 1) data representation; and 2) feature extraction. The former explores how to represent irregular point cloud data by regular grid structures, and the latter continues to explore how to efficiently and elegantly extract features in the transformed regular representation domain. Architecturally, our method shares similar big pictures with previous rasterization-based learning frameworks in terms of creating regular representations for irregular geometric data. However, volumetric grids and multi-view images are irreversible representations, which means that it is basically impossible to accurately recover the original point clouds, and hence are typically restricted as pre-processing steps to support regular-domain processing techniques. Differently, PGI serves as a generic representation modality for 3D point clouds, which is highly accurate and naturally reversible. In a sense, our job is simply to rearrange unordered points with learned ``canonical'' orders without destroying the original geometry information.

In summary, this work makes the following key contributions:
\begin{itemize}
	\item \mr{we represent irregular 3D point clouds by regular 2D PGIs via Flattening-Net, which can be conveniently implemented in an unsupervised manner and turns to be transferable between different data domains;}
	\item \mr{we customize CSConv for unified, scalable, and efficient surface-style point feature extraction on PGIs; and}
	\item \mr{we design task-specific networks for rich application scenarios to enable practical verification of the potential of our PGI representation structures.}
\end{itemize}

\mr{The remainder of this paper is organized as follows. In Section \ref{related-work}, we discuss three families of closely-related works and include brief reviews of specialized deep learning-based 3D point cloud processing fields involved in our subsequent experiments. Section \ref{sec:flattening-net} introduces the proposed Flattening-Net for generating regular PGI representation structures from irregular point clouds in an unsupervised manner. Section \ref{sec:feature-learning} further proposes CSConv that directly operates on PGIs for surface-style regional  embedding and builds different task-specific networks. Section \ref{sec:experiments} begins with quantitative experimental evaluations of PGI representation quality, and then provides downstream task performances and comparisons. In Section \ref{sec:diff-pgi-gi}, we additionally present in-depth discussions about some critical issues to help better assess and understand the value of our work. Finally, we conclude this paper in Section \ref{sec:conclusion}.}

\section{Related Work} \label{related-work}

In this section, we mainly focus on three aspects of research advancement closely related to the actual scope of our work: parameterization-, surface-, and deformation-based models for 3D geometric signal learning. Additionally, considering that our proposed PGI representation structure is applied to a variety of downstream point cloud processing applications for experimental evaluation, we also include necessary discussions about representative task-driven deep learning frameworks for completeness.

\subsection{Parameterization-based Models} \label{sec-2-1}
In the geometry processing community, there exist a family of \textit{parameterization-based} works that are devoted to generalizing standard CNNs in regular domains to geometric deep learning architectures in irregular domains. Masci \textit{et al}. \cite{masci2015geodesic} proposed to construct a polar coordinate system to extract local 3D patches and defined geodesic convolutions on manifolds. Later, Boscaini \textit{et al}. \cite{boscaini2016learning} used oriented anisotropic diffusion kernels as an alternative patch operator. Monti \textit{et al}. \cite{monti2017geometric} employed Gaussian mixture kernels to implement a parametric, instead of fixed, patch extraction procedure. These methods parameterize local 3D surfaces around each query point and formulate convolution as intrinsic template matching, which cannot fully incorporate global context. To remedy this issue, Sinha \textit{et al}. \cite{sinha2016deep} adopted global spherical parameterization to create geometry images (GIs), as initialized in \cite{gu2002geometry}, from genus-zero manifolds. The follow-up work \cite{sinha2017surfnet} further investigated a correspondence-based procedure to produce consistent GIs for shapes of the same category. Maron \textit{et al}. \cite{maron2017convolutional} developed a global seamless parameterization method that maps sphere-like surfaces to a flat torus. To generate low-distortion surface-to-image representations, Haim \textit{et al}. \cite{haim2019surface} employed a covering map and provided a learning framework for spherical signals. These methods generate regular representations of 3D geometry, which make it possible to introduce standard CNNs without modifications. Unfortunately, their frameworks are still built upon traditional optimization algorithms and only apply to sphere-type (genus-zero) surfaces. Moreover, these methods operate on mesh models with connectivity information and thus are not directly applicable to unstructured point clouds. In particular, a more generic regular geometry parameterization scheme can be found in \cite{ezuz2017gwcnn}, which supports a wider range of geometric modality like meshes and point clouds.

\subsection{Surfaced-based Models} \label{sec-2-2}
In real-world applications, 3D geometric signals are usually generated by various depth sensors and LiDAR scanners to describe the actual boundary surfaces of physical objects and scenes, making ``surface'' a natural and critical representation modality for 3D geometry. Accordingly, there exist a special family of \textit{surface-based} models that directly work on surface geometry of point clouds and are supposed to be less vulnerable to surface deformations. Tatarchenko \textit{et al}. \cite{tatarchenko2018tangent} proposed to project local neighbors of each spatial point onto the corresponding tangent plane and explored different interpolation schemes to construct virtual tangent images for performing convolutions. However, this method is sensitive to normal estimation. Lin \textit{et al}. \cite{lin2020fpconv} extended the projection-interpolation working mechanism to a more general and implicit learning process by predicting the soft weighting maps, which can be more flexible and robust to deal with complex shapes. Komarichev \textit{et al}. \cite{komarichev2019cnn} presented a new orientation-invariant annular convolution operator that defines convolutional kernels in ring-shaped regions on the projection domain. This method directly learn features from points without interpolation, but still needs to approximate point normals through explicit plane fitting. Comparatively, in our method, there is no such need to estimate normals or to fit planes which are cumbersome and vulnerable, and the original geometry in the 3D space is preserved. \mr{In addition to the above plane-based learning strategies, other methods \cite{cao20173d,esteves2018learning,coors2018spherenet,cohen2018spherical,kondor2018clebsch,jiang2018spherical,cohen2019gauge,rao2019spherical} also investigated an alternative paradigm of performing convolution on spherical signals. Coors \textit{et al}. \cite{coors2018spherenet} adapted convolutional filters by encoding projection distortions. Jiang \textit{et al}. \cite{jiang2018spherical} proposed to define spherical convolutions as a learned linear combination of parameterized differential operators. Other different line of works \cite{cohen2018spherical,esteves2018learning,kondor2018clebsch} particularly focus on learning rotation-invariant representations from spherical signals.}

\subsection{Deformation-based Models} \label{sec-2-3}
Based on the observation that point clouds are discretized or sampled from object surfaces, researchers have developed a series of \textit{deformation-based} neural architectures used for 3D shape generation, point cloud reconstruction, and unsupervised learning. Yang \textit{et al}. \cite{yang2018foldingnet} proposed to deform an initial 2D lattice grids to reconstruct input point clouds through a folding-based decoder that is built upon shared MLPs, in an attempt to enhance the representation ability of intermediate feature encodings. Groueix \textit{et al}. \cite{groueix2018papier} used multiple learnable parameterizations to decode the target surface. Considering that genus-zero primitives (\textit{e.g.}, lattice grids) are insufficient to capture highly complex topological structures, Chen \textit{et al}. \cite{chen2019deep} presented a graph-based decoder with graph topology inference and filtering. Instead of using manually specified uniform grid structures, Deprelle \textit{et al}. \cite{deprelle2019learning} attempted to use learned 3D templates that are pre-generated from training data. Recently, Pang \textit{et al}. \cite{pang2021tearingnet} presented a topology-friendly auto-encoder that enables to adaptively break the edges of a single primitive graph, making it easier to tackle objects of various genera or scenes with multiple components. The core intuition of these methods is that neural networks tend to fit a relatively smooth transformation function, such that adjacent data points tend to be mapped to close locations in the target space. In summary, these methods deform initial templates to target shapes, differently, our method attempts to flatten 3D surface points onto 2D planar domains, which turns to be the exactly opposite working direction (3D-to-2D \textit{vs.} 2D-to-3D) with complementary characteristics.

\subsection{Deep Learning-based Point Cloud Processing} \label{sec-2-4}
In recent years, deep learning has become the dominating technique for almost all mainstream point cloud processing and understanding tasks. In addition to shape classification \cite{wu20153d} and part segmentation \cite{yi2016scalable} that serve as the most popular tasks for benchmarking generic point cloud networks, here we restrict our attentions to point cloud reconstruction and upsampling application scenarios. We refer the readers to \cite{guo2020deep} for a comprehensive survey on deep learning for 3D point clouds.

Driven by reconstruction objectives, auto-encoders play a crucial role in representation learning of point clouds, with rich applications in 3D surface modeling, shape editing, and unsupervised learning. Wu \textit{et al}. \cite{wu2016learning} extended generative adversarial networks (GANs) from 2D images to 3D voxels. Li \textit{et al}. \cite{li2017grass} introduced a recursive auto-encoder to capture the hierarchical organization of object parts. Achlioptas \textit{et al}. \cite{achlioptas2018learning} built a point cloud auto-encoder upon MLPs and then trained a plain GAN in the fixed latent
space. Deformation-based networks \cite{yang2018foldingnet, groueix2018papier}, as discussed in Section \ref{sec-2-3}, also demonstrated great potential in generating compact shape encodings. Implicit models \cite{park2019deepsdf, mescheder2019occupancy, chen2019learning} aim to describe continuous surfaces without discretization by implicit fields, which can be more efficient for high-quality representation.

Following 2D image super-resolution \cite{dong2015image, kim2016accurate}, 3D point cloud upsampling is also attracting growing attentions. The pioneering deep learning architecture PU-Net \cite{yu2018pu} learned multi-scale point-wise features and expanded sparse point sets through multi-branch MLPs. Yu \textit{et al}. \cite{yu2018pu} presented a multi-step progressive upsampling (MPU) framework for capturing and restoring different levels of geometric details, which can be computationally expensive due to multi-stage supervision. Li \textit{et al}. \cite{li2019pu} introduced GANs to construct PU-GAN for points distribution modeling, in order to improve the uniformity of the upsampling results, which works well on non-uniform point cloud data. Qian \textit{et al}. \cite{qian2020pugeo} presented PUGeo-Net, a geometry-centric upsampling framework that generates dense samples in 2D parametric domains, which are further lifted to the 3D space through linear transformations to touch the target surface.

\begin{figure*}[t]
	\centering
	\subfigure[Grid-to-Surface Deformation (G2SD)]{\includegraphics[width=1.0\linewidth]{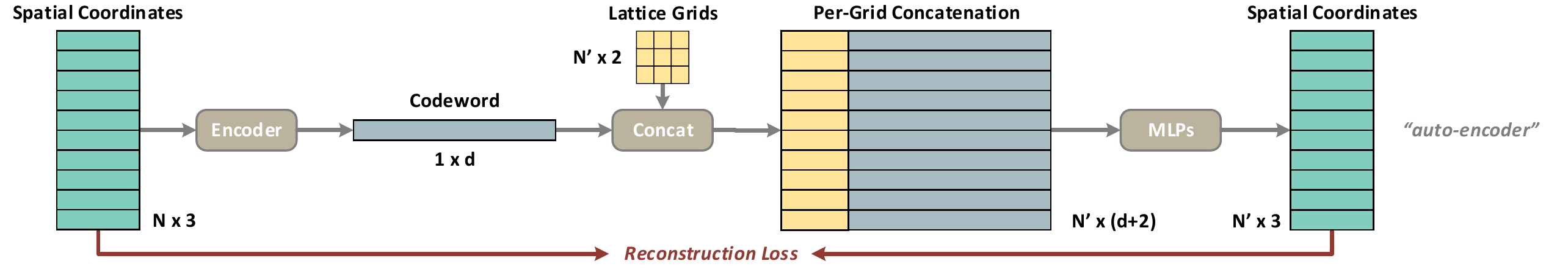} }
	\subfigure[Surface-to-Plane Flattening (S2PF)]{\includegraphics[width=1.0\linewidth]{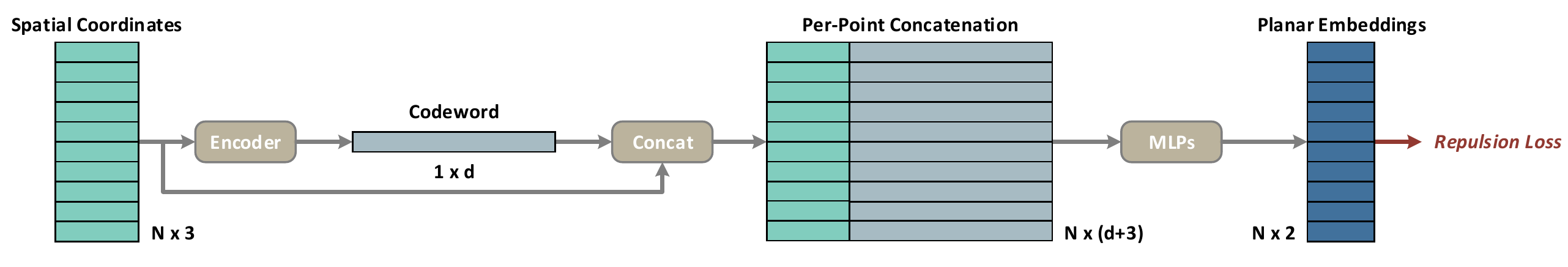}}
	\caption{Overall workflows of G2SD and S2PF in terms of learning point-wise mappings between 2D and 3D domains. (a) The G2SD architecture, formulated as an auto-encoder, tends to deform fixed 2D lattice grids to reconstruct the input 3D point cloud. (b) The S2PF architecture solves the opposite problem, which explicitly maps the input 3D points onto a 2D plane driven by a repulsion loss.}
	\label{fig:comparison-g2sd-s2pf}
\end{figure*}

\section{Flattening-Net for PGI Generation} \label{sec:flattening-net}

\subsection{Overview} \label{overview}
Given an input 3D point set $\mathcal{P} \in \mathbb{R}^{N \times 3}$ containing $N$ points, Flattening-Net is designed to convert $\mathcal{P}$ into a regular 2D PGI representation structure $\mathcal{I} \in \mathbb{R}^{3 \times m \times m}$. Here, $m$ is a user-specified factor that determines the image resolution, and thus $M = m \times m$ denotes the number of pixels in the generated PGI. We can directly obtain the corresponding 3D point set $\mathcal{Q} \in \mathbb{R}^{M \times 3}$ encoded in its equivalent image form $\mathcal{I}$ through reshaping operations. 
 
Technically, Flattening-Net is a two-stage learning architecture consisting of a surface flattening module (SFM) that can map spatial points to planar embeddings hierarchically, such that each 2D point on the embedded plane corresponds to a certain 3D point on the input shape, and a grid resampling module (GRM) that redistributes embedding points on a uniform lattice to construct a regular grid structure. In what follows, we detail these core components one by one.

\subsection{Surface Flattening Module} \label{SFM}

In this module, our goal is to build \mr{a locally smooth mapping} between 3D object surfaces and 2D planar domains through point-wise embedding. To achieve this, we investigate two learning architectures for parameterizing irregular 3D point clouds onto 2D lattice grids: 1) Grid-to-Surface Deformation (G2SD); 2) Surface-to-Plane Flattening (S2PF). Conceptually, G2SD deforms a pre-defined 2D lattice to the target 3D point cloud, while S2PF adopts an opposite workflow that maps 3D spatial points to 2D planar embeddings.

Despite the architectural similarities, these two learning approaches have complementary characteristics in terms of generating planar flattenings, \textit{i.e.}, G2SD is good at producing locally smooth parameterizations for complete objects, while S2PF is able to flatten local surfaces in a geometrically meaningful manner. However, they both suffer from similar computational limitations in processing dense point clouds and cannot deal with complex topological structures. These observations motivate us to formulate a hierarchical hybrid framework to achieve high-quality regular geometry parameterizations while maintaining efficiency. 

In the following, we first introduce the detailed design of G2SD and S2PF in Section \ref{3-2-1} and \ref{3-2-2}, and then provide experimental comparisons to explain their complementarity, based on which we describe how the hierarchical hybrid framework is constructed and trained in Section \ref{3-2-3}.

\subsubsection{Grid-to-Surface Deformation (G2SD)} \label{3-2-1}

The basic idea of G2SD has been explored in existing works \cite{yang2018foldingnet, groueix2018papier} that deform fixed 2D lattice grids to form a target 3D shape. In these works, this \textit{2D-to-3D} shape deformation process is particularly developed for 3D object generation or unsupervised learning under an auto-encoder framework, with purposes of enhancing the representation ability of the intermediate global shape encodings.
Differently, 
we re-consider G2SD as a generative model to achieve regular geometry parameterization, \textit{i.e.}, an indirect way of flattening 3D surface geometry onto the 2D lattice space. 

As shown in Figure \ref{fig:comparison-g2sd-s2pf},  both G2SD and S2PF begin with extracting a 1D codeword $\mathbf{z}$ from input points to encode the global shape information compactly, which can be achieved by any commonly-used deep set encoders. In our implementation, we adopt PointNet-vanilla \cite{qi2017pointnet} to obtain point-wise embeddings and perform max-pooling to obtain $\mathbf{z}$.

After codeword extraction, the subsequent workflow of the G2SD architecture can be formulated as
\begin{equation} \label{eq-01}
	\mathcal{R}=\phi([\mathcal{D}; \mathbf{z}]),
\end{equation}
where $\mathcal{D} \in \mathbb{R}^{N^{\prime} \times 2}$ defines a 2D lattice with $n^{\prime} \times n^{\prime}$ uniformly distributed grids ($N^{\prime} = n^{\prime} \times n^{\prime}$), $[\cdot;\cdot]$ represents channel concatenation, and $\phi(\cdot)$ denotes a non-linear transformation operator that can be implemented by shared MLPs. In such an auto-encoding framework, we reconstruct a point cloud $\mathcal{R} \in \mathbb{R}^{N^{\prime} \times 3}$ that is supposed to recover the input point cloud $\mathcal{P} \in \mathbb{R}^{N \times 3}$ under some point-set similarity metrics, such as Chamfer Distance (CD) and Earth Mover's Distance (EMD). To deal with complex 3D shapes, we can successively stack multiple G2SD units to empower the whole reconstruction framework. For example, a two-stack learning architecture can be described as $\mathcal{R}=\phi_2([\phi_1([\mathcal{D}; \mathbf{z}]); \mathbf{z}])$, where $\phi_1(\cdot)$ and $\phi_2(\cdot)$ denote two separate layers of shared MLPs.

Intuitively, since neural networks tend to learn a relatively smooth transformation function, we can always expect that neighboring grids on the 2D lattice are locally utilized as a whole to approximate a 3D patch region on the target shape. Thus, we can geometrically interpret G2SD as ``paper-folding''.

\subsubsection{Surface-to-Plane Flattening (S2PF)} \label{3-2-2}
In contrast to G2SD, we propose a new working mechanism called S2PF to explore the possibility of inversely flattening 3D shapes onto 2D planes. The proposed S2PF shares some spirit of G2SD, as they both adopt shared MLPs as building blocks to obtain point-wise mappings between 3D and 2D domains. Following the preceding notations, we formulate our S2PF architecture as
\begin{equation} \label{eq-02}
	\mathcal{F}=\sigma(\phi([\mathcal{P}; \mathbf{z}])),
\end{equation}
where $\sigma(\cdot)$ represents the sigmoid function used to restrict the generated planar embedding points $\mathcal{F} \in \mathbb{R}^{N \times 2}$ within a unit square domain $[0, 1]^2$.

As we discussed previously, since $\phi(\cdot)$ approximates a smooth mapping, points from the original surface can be flattened in a locally-continuous manner. For convenience, we denote the $i$-th spatial point in $\mathcal{P}$ and its embedding point in $\mathcal{F}$ as $\mathbf{p}_i=(x_i, y_i, z_i)$ and $\mathbf{f}_i=(u_i, v_i)$. For any two neighboring embedding points $\mathbf{f}_i$ and $\mathbf{f}_j$, we can expect that their corresponding pre-images $\mathbf{p}_i$ and $\mathbf{p}_j$ should also be close to each other in the original 3D space.

Different from G2SD that is optimized by reconstruction loss functions, we propose to impose a repulsion constraint to stretch point distribution by punishing clustered pairs of planar embedding points until they can be separated by an appropriate distance threshold $\epsilon$. Mathematically, we define the repulsion constraint on the generated $\mathcal{F}$ as
\begin{equation} \label{eq-03}
	\mathcal{L}_{\mathrm{repulsion}}(\mathcal{F}) = \sum_{i=1}^{N} \max\left(0, \epsilon - \left\| \mathbf{f}_i-\mathbf{f}_j \right\|_2\right),
\end{equation}
where $\left\| \cdot \right\|_2$ computes the $L_2$ norm of a vector, and $\mathbf{f}_j$ is the unique nearest neighbor of $\mathbf{f}_i$ in the embedding point set $\mathcal{F}$. We set the distance threshold $\epsilon=1/(m-1)$, which means the grid interval of a uniform $m \times m$ lattice and works well. In practice, we observe that the learning process is basically insensitive to a smaller value of $\epsilon$, since it does not hurt the separability of embedding points and we always perform normalization to re-scale $\mathcal{F}$ into $[0, 1]^2$.

\begin{figure}[t]
	\centering
	\includegraphics[width=1.0\linewidth]{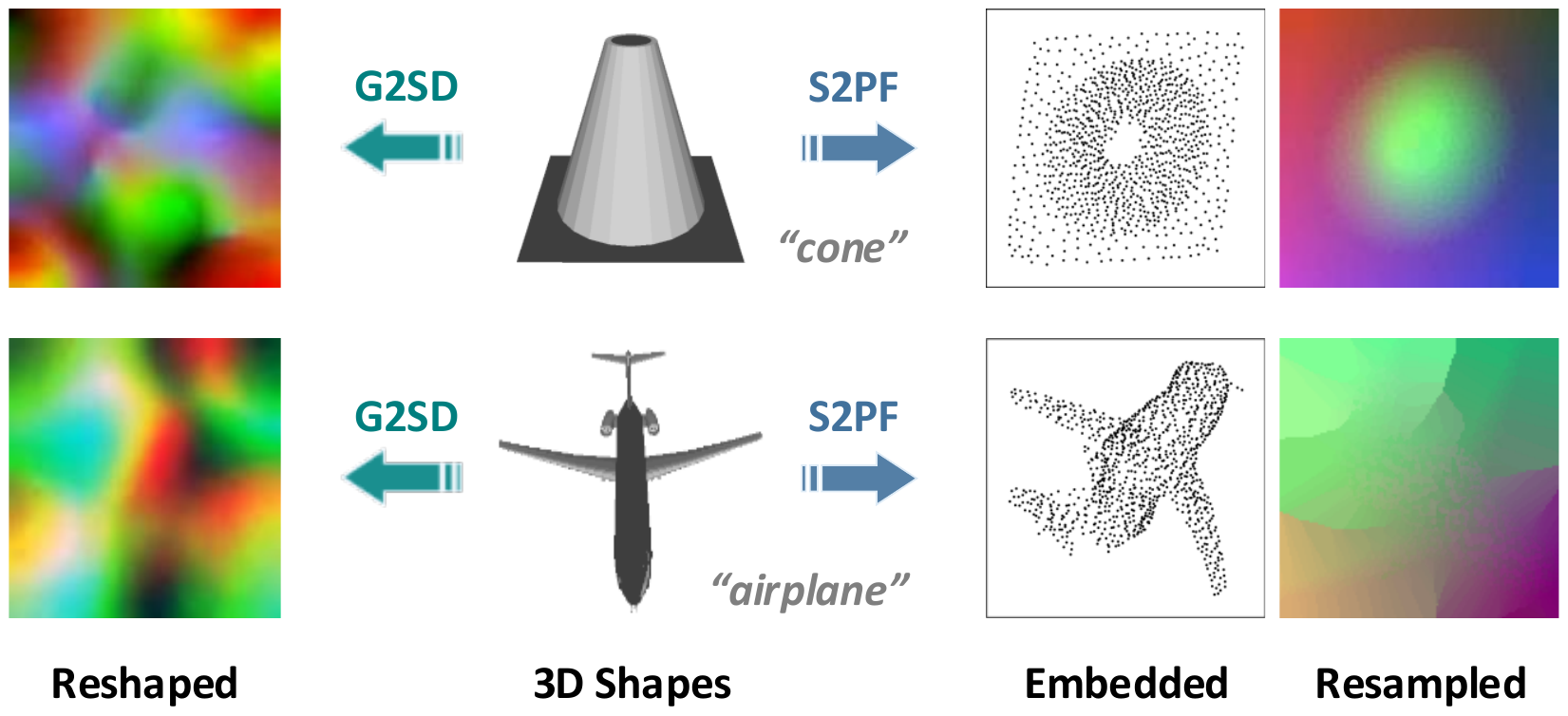}
	\caption{Comparison of G2SD and S2PF in terms of regular geometry parameterization. Given a target shape represented by $N$ spatial points, G2SD reconstructs an $N^{\prime} \times 3$ point cloud that is further reshaped as a $3 \times n^{\prime} \times n^{\prime}$ 2D image, S2PF directly maps input points onto a 2D plane and then performs resampling on lattice grids to construct a 2D image. We treat point coordinates as pixel colors to visualize the corresponding three-channel images.}
	\label{fig:complementarity-g2sd-s2pf}
\end{figure}

\textbf{Comparisons between G2SD and S2PF}. The G2SD and S2PF learning architectures show complementary features when applied to generate regular geometry representations, and also share obvious limitations in some practical aspects.

Through extensive experiments, we observe that G2SD is always able to produce locally smooth parameterizations, even for complex topological structures, but fails to unfold 3D surfaces in a geometrically-meaningful manner. By contrast, S2PF can ``flatten'' local surfaces or objects with simple topology in a true sense. Figure \ref{fig:complementarity-g2sd-s2pf} shows two typical examples. For the simple \textit{cone} object, G2SD simply preserves local smoothness, while S2PF flattens it with global continuity. However, for the complex \textit{airplane} object, S2PF cannot find a nice ``cut'' to ``open up'' the complete object, but tends to learn a ``view projection'', which deviates from our objective to maintain local neighborhood consistency. 

In addition, it turns out that both G2SD and S2PF cannot deal with dense point clouds, which limits their potentials in capturing geometric details. As the number of consumed points increases, the training cost (including GPU memory footprint and computational complexity) grows fast, and it can be much harder to converge.

Based on these observations, we are motivated to design a hierarchical hybrid flattening architecture, as detailed in the following subsection, such that we can produce high-quality flattenings for complex shapes while maintaining efficiency when dealing with dense inputs.

\subsubsection{Hierarchical Hybrid Flattening} \label{3-2-3}
Our basic idea is to decompose the input point cloud into a sparse set of guidance points and the corresponding context points, which are separately flattened onto planar domains and then assembled to form a complete geometry representation in an image-like structure. Intuitively, we treat G2SD as a ``teacher'' network to guide the learning of its ``student'' S2PF network, which fully exploits the complementarity of the two opposite learning architectures.

Technically, we start by applying farthest point sampling (FPS) to generate guidance points $\mathcal{P}_G \in \mathbb{R}^{N_G \times 3}$ from input points $\mathcal{P} \in \mathbb{R}^{N \times 3}$. Treating each guidance point as a patch centroid, we employ $k$-NN to search $N_C$ spatial neighbors, which can deduce a collection of context points $\mathcal{C} = \{\mathcal{C}_i\}_{i=1}^{N_G}$, where $\mathcal{C}_i \in \mathbb{R}^{N_C \times 3}$.

To achieve hierarchical flattening, we first train a G2SD network that consumes guidance points $\mathcal{P}_G$ and pre-defined lattice grids $\mathcal{G} \in \mathbb{R}^{2 \times n_G \times n_G}$ to reconstruct $N_G = n_G \times n_G$ points, which are reshaped into $\hat{\mathcal{P}}_G \in \mathbb{R}^{3 \times n_G \times n_G}$. We denote the entries of $\mathcal{G}$ and $\hat{\mathcal{P}}_G$ at pixel position $(\hat{u},\hat{v})$ as $\mathcal{G}_{(\hat{u},\hat{v})} \in \mathbb{R}^2$ and $\hat{\mathcal{P}}_{G_{(\hat{u},\hat{v})}} \in \mathbb{R}^3$, respectively. According to the working mechanism of G2SD, we know that $\hat{\mathcal{P}}_{G_{(\hat{u},\hat{v})}}$ is deformed from fixed $\mathcal{G}_{(\hat{u},\hat{v})}$. Inversely, we can interpret such a mapping relationship as: the spatial point $\hat{\mathcal{P}}_{G_{(\hat{u},\hat{v})}}$ should be flattened onto a 2D grid position which we denote as $\hat{\textbf{f}}=(\hat{u},\hat{v})$. By considering all point pairs between $\mathcal{G}$ and $\mathcal{P}_G$, we can obtain a set of planar embedding points $\hat{\mathcal{F}}_G \in \mathbb{R}^{N_G \times 2}$. Next, we deploy an S2PF network that directly maps $\mathcal{P}_G$ to planar embeddings $\mathcal{F}_G \in \mathbb{R}^{N_G \times 2}$. Here, we consider $\hat{\mathcal{F}}_G$ as preliminary knowledge acquired from the teacher G2SD network, and train the student S2PF network by minimizing $L_1$ loss between $\mathcal{F}_G$ and $\hat{\mathcal{F}}_G$: 
\begin{equation} \label{eq-04}
	\mathcal{L}_{\mathrm{guidance}}(\mathcal{F}_G;\hat{\mathcal{F}}_G) = \sum_{i=1}^{N_G} \left\| \mathbf{f}_{G_i}-\hat{\mathbf{f}}_{G_i} \right\|_1,
\end{equation}
where $\mathbf{f}_{G_i}$ and $\hat{\mathbf{f}}_{G_i}$ represent the $i$-th point in $\mathcal{F}_G$ and $\hat{\mathcal{F}}_G$.

Conclusively, we take three steps to train an S2PF network that flattens guidance points $\mathcal{P}_G$ to $\mathcal{F}_G$ on a 2D plane: 
\begin{enumerate}
    \item train a teacher G2SD network to obtain $\hat{\mathcal{F}}_G$;
    \item initialize a student S2PF network by minimizing $\mathcal{L}_{\mathrm{guidance}}(\mathcal{F}_G;\hat{\mathcal{F}}_G)$; and 
    \item fine-tune the S2PF network under the constraint of $\mathcal{L}_{\mathrm{repulsion}}(\mathcal{F}_G)$.
\end{enumerate}

After that, we deploy another separate S2PF network for flattening context points $\mathcal{C}_i$ to $\mathcal{F}_{C_i} \in \mathbb{R}^{N_C \times 2}$ under repulsion constraints $\mathcal{L}_{\mathrm{repulsion}}(\mathcal{F}_{C_i})$, $i={1,...,N_G}$. Since local patches have simple topological structures, S2PF can easily generate geometrically-meaningful flattenings. Besides, $\mathcal{P}_G$ is sparse and only depicts coarse geometry, which can make it easier for G2SD to generate smoother and more accurate reconstructions.

\begin{figure}[t]
	\centering
	\includegraphics[width=1.0\linewidth]{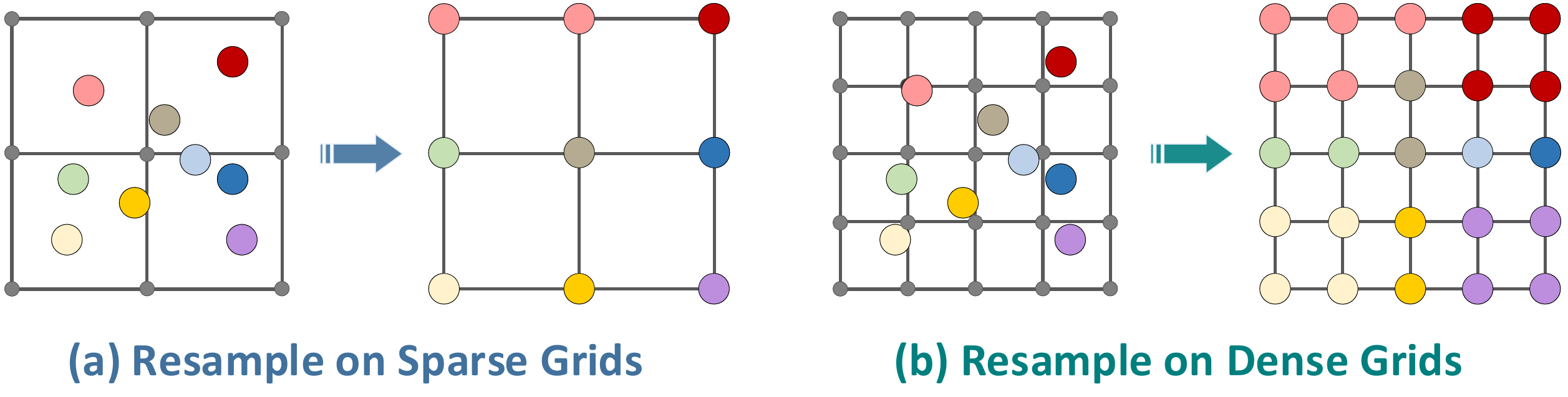}
	\caption{In grid resampling, we map irregular planar embeddings to regular pixels. In general, resampling with sparse grids may yield significant information loss (a), while resampling with dense grids can effectively reduce such loss (b). }
	\label{fig:comparison-sparse-dense-grid-resamping}
\end{figure}

\subsection{Grid Resampling Module (GRM)} \label{GRM}
The preceding SFM embeds the guidance points $\mathcal{P}_G$ and the corresponding context points $\{\mathcal{C}_i\}_{i=1}^{N_G}$ onto the unit square domains, where we obtain $\mathcal{F}_G$ and $\{\mathcal{F}_{C_i}\}_{i=1}^{N_G}$, respectively. GRM aims to assemble them into a complete embedding set, and generate the required regular geometry representation structure by grid resampling.

We consider every single point $\mathbf{f}_{G_i}$ in $\mathcal{F}_G$ as the global mapping coordinate of the local patch $\mathcal{C}_i$, and within $\mathcal{C}_i$ we treat $\mathcal{F}_{C_i}$ as local mapping coordinates for the patch points. However, planar embedding points obtained from SFM are not guaranteed to be located at grid positions of a regular lattice. To this end, we need an additional grid resampling procedure to redistribute the irregular embedding set over uniform grids and generate a completely regular PGI.

Mathematically, we can formulate the GRM as an assignment problem
\begin{equation} \label{eq-05}
\min_{\mathcal{W}^g, \mathcal{W}^l_i} \left\| \mathcal{W}^g \mathcal{G}^g - \mathcal{F}_G\right\|_F + \sum_{i=1}^{N_G} \left\| \mathcal{W}^l_i \mathcal{G}^l_i - \mathcal{F}_{C_i}\right\|_F,
\end{equation}
where we aim to optimize a set of permutation matrices $\mathcal{W}^g \in \mathbb{R}^{N_G \times N_G}$ and $\{\mathcal{W}^l_i\}_{i=1}^{N_G} \in \mathbb{R}^{K \times N_C}$ to uniquely assign the global and local embedding coordinates $\mathcal{F}_G$ and $\{\mathcal{F}_{C_i}\}_{i=1}^{N_G}$ to pre-defined 2D grid points with minimal cost of total movement. Note that $\mathcal{G}^g \in \mathbb{R}^{N_G \times 2}$ denotes an $n_G \times n_G$ lattice such that $\mathcal{W}^g$ defines a bijection with respect to the guidance points $\mathcal{F}_G$, while $\mathcal{G}^l_i \in \mathbb{R}^{K \times 2}$ denotes a redundant $k \times k$ lattice with $K = k \times k > N_C$, which means that only a subset of the grid points form a one-to-one correspondence with the context points $\mathcal{F}_{C_i}$. We fill in the rest unmatched grid positions in $\mathcal{G}^l_i$ by selecting the closest neighbors from $\mathcal{F}_{C_i}$. As illustrated in Figure \ref{fig:comparison-sparse-dense-grid-resamping}, using denser grids inevitably introduces representation redundancy, but it can effectively reduce loss of points during resampling. In practice, Eq. (\ref{eq-05}) can be efficiently solved by the Auction algorithm \cite{bertsekas1990auction}.

Combining global and local mapping coordinates after grid assignment, we can obtain a $k \times k$ square block, which we call a geometry image block, for each patch of context points, after which we globally assemble all blocks into a complete PGI denoted as $\mathcal{I} \in \mathbb{R}^{3 \times m \times m}$. Obviously, a PGI is composed of $n_G \times n_G$ square blocks, each of which contains $k \times k$ pixels. Following the preceding notations, we have $M = N_G \times K$ and $m = n_G \times k$. \mr{For simplicity, we uniformly configure $n_G=16$ in all experimental setups.}

\section{Deep Feature Learning from PGIs} \label{sec:feature-learning}

\mr{The preceding section introduces Flattening-Net for creating regular PGI representation structures from raw point clouds. In this section, we will take a step forward towards learning deep features directly from the generated PGIs, after which we can equivalently achieve various point cloud processing and understanding tasks.}

\mr{As afore-mentioned, since the overall surface embedding process is implemented through a hierarchical (global-local) flattening workflow, PGIs are intrinsically block-structured. Besides, local patch parameterizations (obtained from S2PF) within the blocks are produced in a geometrically meaningful manner, which naturally supports surface-style feature aggregation paradigms. Hence, motivated by such two aspects of properties, we customize a novel concentric-square convolution (CSConv) operator, as introduced in Section \ref{sec:csconv}, based on which we further construct a unified regional embedding layer that operates on PGIs to efficiently extract local geometry descriptors from all the $n_G \times n_G$ blocks. To enable practical evaluations on actual applications, in Section \ref{sec:task-networks}, we make additional efforts on task-specific network design for tackling diverse types of downstream point cloud processing and understanding scenarios, including high-level tasks of classification and segmentation and low-level tasks of reconstruction and upsampling. Functionally, the proposed CSConv is designed to serve as a plug-in component in the whole processing pipeline to bridge the front-end Flattening-Net and the back-end task-specific network, with purpose of achieving efficient and scalable downstream learning while in the meantime sufficiently exploiting the unique structural properties of PGIs.}

\mr{In the following, we comb the core working mechanism and sketch the major architectural design for readability and brevity. We also refer readers to the supplementary material and source code for more detailed and complete technical implementations.}

\subsection{CSConv for Regional Embedding} \label{sec:csconv}

\begin{figure}[t]
	\centering
	\includegraphics[width=0.85\linewidth]{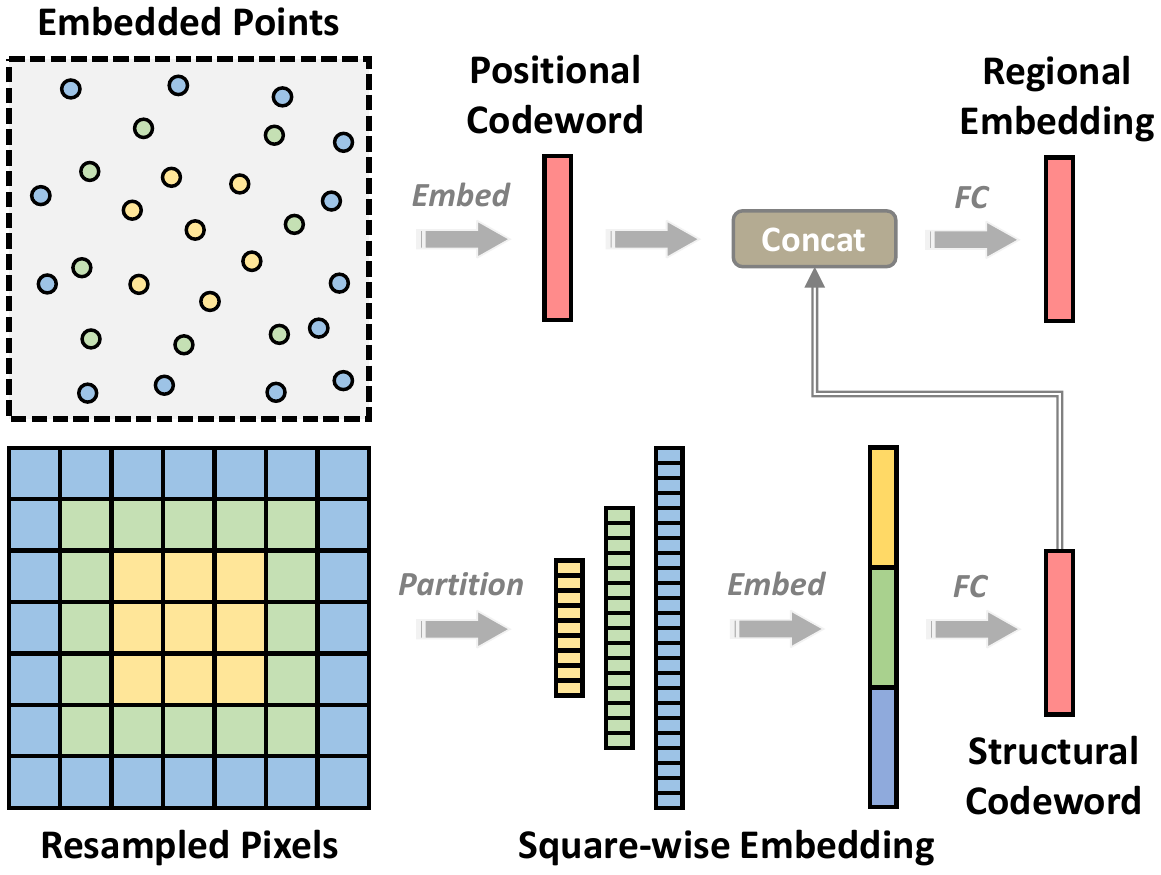}
	\caption{\mr{Illustration of the proposed CSConv operator directly working on PGI representation structures to achieve efficient and scalable regional embedding. Given a geometry image block whose pixels are resampled from embedded spatial points, we sequentially partition the whole block scope into innermost-, intermediate-, and outermost-squares. Treating each square as a point set, we adopt shared MLPs followed by channel-wise max-pooling to output vectorized square-wise embeddings, which are concatenated in order and further fused through a separate FC layer into a structural codeword. In parallel, we perform absolute coordinates embedding to generate a positional codeword. Finally, we concatenate and fuse the structural and positional codewords to obtain the regional embedding vector.}}
	\label{fig:illustration-of-csconv}
\end{figure}

\mr{As depicted in Figure \ref{fig:illustration-of-csconv}, our core motivation lies in modeling 3D surface geometry by means of an ordered (from central to peripheral scope) sequence of concentric squares partitioned separately from each geometry image block.}

\mr{More specifically, we consider a square-shaped geometry image block, denoted as $\mathcal{B} \in \mathbb{R}^{3 \times k \times k}$, containing $K$ points. We partition the whole block into three concentric squares, \textit{i.e.,} innermost, intermediate, and outermost square regions, sequentially denoted as $\{ \mathcal{S}_{\mathrm{inner}}, \mathcal{S}_{\mathrm{inter}}, \mathcal{S}_{\mathrm{outer}} \}$. Treating each concentric square as a separate point set, we pre-normalize each set of points into a unit sphere for capturing structural information from the relative coordinates, and deploy shared MLPs to generate high-dimensional point-wise embeddings. We perform channel-wise max-pooling to deduce, from each of the three concentric square regions, the corresponding vectorized representations $\{ \mathbf{v}_{\mathrm{inner}}, \mathbf{v}_{\mathrm{inter}}, \mathbf{v}_{\mathrm{outer}} \}$, which are concatenated in order and further fed into a fully-connected (FC) layer for inter-squares fusion, producing a structural codeword $\mathbf{v}_r$. In parallel, in order to gain awareness of positional information from absolute coordinates of local patch points, we directly deploy $1 \times 1$ convolutions to the image structure of block $\mathcal{B}$ and then apply $k \times k$ spatial max-pooling to produce a positional codeword $\mathbf{v}_a$. Finally, we concatenate the structural codeword $\mathbf{v}_r$ and the positional codeword $\mathbf{v}_a$ to generate a regional embedding vector $\mathbf{v} \in \mathbb{R}^{d_v}$. Hence, given a complete PGI representation structure composed of $n_G \times n_G$ geometry image blocks, we obtain a regular 2D feature map denoted as $\mathcal{V} \in \mathbb{R}^{d_v \times n_G \times n_G}$ through CSConv. In all experimental setups, we uniformly generate fixed-length regional embeddings with $d_v = 128$.}

\mr{Generally, CSConv serves as a customized operator for PGIs, featured by square-wise embedding and aggregation, which facilitates learning the underlying manifold structure of the corresponding local patch surface. By bridging PGIs and task-specific networks via CSConv, the overall downstream processing pipeline uniformly begins with the  low-resolution feature map (\textit{i.e.,} $\mathcal{V}$) assembled by regional embedding vectors, achieving satisfactory efficiency as well as scalability. As the number of input points (\textit{i.e.,} $N$) increases significantly, the computational complexity of CSConv only grows moderately and there is no need to adjust the subsequent network configuration, since practically $n_G$ is usually much smaller than $m$ and can be maintained unchanged for tackling different number of input points.}

\subsection{Task Network Design} \label{sec:task-networks}

\mr{To evaluate the potential and effectiveness of the proposed PGI representation structure, we make additional efforts to design downstream task networks for four typical types of point cloud learning applications, including 1) classification, 2) segmentation, 3) reconstruction, and 4) upsampling. For the first three scenarios, the task-specific network is connected to the preceding CSConv-driven regional embedding component, meaning that the subsequent feature extraction pipeline uniformly starts from a regional embedding feature map (\textit{i.e.,} $\mathcal{V}$). In particular, considering that the processing pipeline of point cloud upsampling separately operates on small 3D patches locally constructed from the whole shape, we directly convert the input patch with $N$ points to its 2D PGI representation structure with a resolution of $n \times n$ (where $N = n^2$) without the hierarchical parameterization strategy. Thanks to the structural conversion, we can regard 3D point cloud upsampling as 2D image super-resolution, which is implemented by standard 2D CNN architectures without introducing CSConv for regional embedding before the subsequent task network. In what follows, we sketch the corresponding task-specific network design one by one.}\\

\noindent \textbf{Task Network for Classification.} \mr{As a global geometry understanding problem, 3D shape classification is perhaps the most common and fundamental benchmark task for the evaluation of point cloud learning models. In general, our goal is to deduce a vectorized global shape signature, which is further transformed to the final category scores through a stack of several FC layers.}

\mr{In our implementation, we investigate two different variants of deep shape classifiers, dubbed as FlatNet-Cls-P and FlatNet-Cls-D, by resorting to two classic and widely used point feature extraction paradigms as proposed in PointNet \cite{qi2017pointnet} and DGCNN \cite{wang2019dynamic}, respectively. For the FlatNet-Cls-P variant, we apply a stack of $1 \times 1$ convolutional layers to $\mathcal{V}$ and obtain a vectorized shape signature by 2D global max-pooling. For the FlatNet-Cls-D variant, we adopt EdgeConv \cite{wang2019dynamic}, a graph-style point convolution operator, to generate point-wise embeddings and then obtain a vectorized shape signature via concatenating the outputs of both 2D global max-pooling and average pooling.}\\

\noindent\textbf{Task Network for Segmentation.} \mr{Instead of categorizing the whole geometric shape, 3D object/scene segmentation is a more fine-grained point cloud understanding task for per-point semantic labeling. Accordingly, our goal is to deduce point-wise features, instead of a single global feature vector, which are then point-wisely mapped to semantic categories through several layers of shared MLPs.}

\mr{Considering the similarity between the classification and segmentation tasks in terms of the overall technical pipeline, here we extend the preceding two variants of classification task networks to the segmentation scenario, namely FlatNet-Seg-P and FlatNet-Seg-D, through replacing the last spatial pooling operations with the top-down feature interpolation procedure as adopted in \cite{qi2017pointnet++}. In particular, for typical scene segmentation tasks where per-point colors are consumed as additional input information, we append color attributes to the corresponding pixel positions to create a $6$-dimensional (\textit{i.e.,} coordinates and colors) image structure to be fed into CSConv for regional embedding. Accordingly, we construct the task network variant called FlatNet-Scene-Seg by resorting to the progressive local feature aggregation mechanism as proposed in \cite{hu2021learning}.}\\

\noindent \textbf{Task Network for Reconstruction.} \mr{Auto-encoders serve as a classic unsupervised learning framework consisting of an encoding process for extracting compact feature representations and a decoding process for reconstructing input signals. Motivated by the structural regularity of PGIs, we implement point cloud reconstruction under an image auto-encoding pipeline, such that we can naturally incorporate standard 2D spatial convolutions.}

\mr{Accordingly, we design a PGI-driven task network called FlatNet-Rec. In the encoding stage, for an input point cloud and its corresponding PGI, we also begin with an $n_G \times n_G$ regional embedding feature map $\mathcal{V}$ passing through a series of convolutional layers accompanied by $2 \times$ spatial pooling. Thus, we compactly deduce a 2D feature map and reshape it into a vectorized global codeword. In the decoding stage, we feed the learned global codeword into another group of convolutional layers accompanied by $2 \times$ spatial up-scaling to generate an $n_G \times n_G$ coarse feature map denoted as $\mathcal{V}_\mathrm{dec}$. On one hand, we aim to restore $N_G$ guidance points from $\mathcal{V}_\mathrm{dec}$, serving as side-output supervision. On the other hand, we locally generate patches centered at each restored guidance point through shared MLPs. A complete reconstruction result is produced by assembling all locally restored patches. The overall training objective involves both pixel-wise L1 loss and point-wise CD loss.}\\

\noindent \textbf{Task Network for Upsampling.} \mr{As afore-mentioned, the current community implements point cloud upsampling via a patch-based processing pipeline, instead of consuming the whole object/scene directly as input. The training process relies on paired data of sparse and dense patches locally constructed from complete 3D models. During the inference phase, an input sparse point cloud is redundantly decomposed into a collection of overlapping patches, which will be separately upsampled. In order to obtain a complete dense point cloud with the explicitly specified scale factor, one can assemble points from all the upsampled patches and then apply FPS to sample the required number of points.}

\mr{Here, our major motivation is to convert the problem of 3D point cloud upsampling to 2D image super-resolution. Architecturally, 2D convolutional learning frameworks are adopted to produce high-resolution PGIs, which are equivalent to dense point sets. Following previous development protocols in \cite{li2019pu}, we experiment with $4 \times$ point upsampling, which corresponds to $2 \times$ image super-resolution. Formally, given a sparse local patch containing $N$ points as well as its PGI representation structure with a resolution of $n \times n$, we aim to generate a larger $2n \times 2n$ PGI as an indirect way of obtaining a denser set of $4N$ points.}

\mr{By resorting to previous successful single image super-resolution frameworks, we construct a task network dubbed as FlatNet-Ups, where we implement spatial interpolation of PGIs under the popular pre-upsampling \cite{dong2015image} and global residual learning \cite{kim2016accurate} paradigm. For an input low-resolution PGI (LR-PGI), we begin with applying the $2 \times$ bicubic image interpolation to compute an enlarged LR-PGI, which further passes through a series of convolutional layers to generate a residual map for restoration of geometric details and outlier removal. Thus, the addition of the enlarged LR-PGI and the residual map gives the desired output of a high-resolution PGI (HR-PGI). The whole learning process is supervised by a combination of pixel-wise L1 loss (between predicted and ground-truth HR-PGIs), point-wise EMD loss, and auxiliary distribution uniformity constraints (as proposed in \cite{li2019pu}).}

\begin{table}[t]
	\renewcommand\arraystretch{1.35}
	\centering
	\caption{Quantitative geometry fidelity metrics computed between the generated PGIs and the original point clouds on ModelNet40 and ShapeNetPart datasets under different number of input points.}
	\setlength{\tabcolsep}{1.5mm}{
		\begin{tabular}{ c | c | c || c }
			\toprule[1.1pt]
			Dataset & \# Points & PGI Resolution & Geometry Fidelity \\
			\hline
			ModelNet40 & $1024$ & $80 \times 80$ & $99.993$\% \\
			ModelNet40 & $2048$ & $112 \times 112$ & $99.915$\% \\
			ModelNet40 & $5000$ & $160 \times 160$ & $99.662$\% \\
			ModelNet40 & $10000$ & $240 \times 240$ & $99.678$\% \\
			ShapeNetPart & $2048$ & $128 \times 128$ & $99.913$\% \\
			\bottomrule[1.1pt]
	\end{tabular}}
	\label{table:pgi-geometry-fidelity}
\end{table}

\begin{table}[t]
	\renewcommand\arraystretch{1.35}
	\centering
	\caption{Trade-off between redundancy and accuracy (geometry fidelity) when generating PGI representation structures of various resolutions from input point clouds uniformly containing 10000 points.}
	\setlength{\tabcolsep}{5.0mm}{
		\begin{tabular}{ c  | c || c }
			\toprule[1.1pt]
			Resolution & Redundancy & Geometry Fidelity \\
			\hline
			$240 \times 240$ & $4.76 \times$ & $99.678$\% \\
			$224 \times 224$ & $4.02 \times$ & $99.532$\% \\
			$208 \times 208$ & $3.33 \times$ & $99.174$\% \\
			$192 \times 192$ & $2.69 \times$ & $98.329$\% \\
			$176 \times 176$ & $2.10 \times$ & $96.688$\% \\
			$160 \times 160$ & $1.56 \times$ & $93.629$\% \\
			$144 \times 144$ & $1.07 \times$ & $88.735$\% \\
			$128 \times 128$ & $0.64 \times$ & $81.537$\% \\
			\bottomrule[1.1pt]
	\end{tabular}}
	\label{table:pgi-trade-off}
\end{table}

\section{Experiments} \label{sec:experiments}

\mr{Generally, we evaluate the potential of our regular geometry representation approach from two perspectives.}
\mr{First, we customized two aspects of computational metrics in terms of geometry fidelity and neighborhood consistency to quantitatively reflect the representation quality of the generated PGIs.}
\mr{Then, we showed the practical effectiveness of FlatteningNet when coupled with the subsequent feature learning pipelines and the corresponding task-specific networks in different downstream applications.}

\subsection{Representation Quality} \label{sec:5.1}

\mr{The proposed PGI representation structure is designed to be a generic geometry modality for point cloud data. Thus, one critical problem is to quantitatively depict the representation quality of the generated PGIs in regular 2D planar domains with respect to the original raw point clouds in irregular 3D domains. Specifically, we customized the following two aspects of computational metrics.}

\textbf{1) Geometry Fidelity.} \mr{We consider an input point cloud $\mathcal{P}$ containing $N$ points and its corresponding PGI representation structure $\mathcal{I}$ of dimensions $m \times m$, which can be equivalently regarded as $M$ points. Considering that the actual information loss is caused by missing points when performing grid resampling, we define the geometry fidelity metric as the ratio of the number of non-missing points to the number of input points (\textit{i.e.,} $N$).   } 

\mr{We performed evaluation on varying resolutions of PGIs, which correspond to varying number of input points, on the whole  ModelNet40 \cite{wu20153d} and ShapeNetPart \cite{yi2016scalable} repositories consisting of 12311 and 16881 3D object models, respectively. As reported in Table \ref{table:pgi-geometry-fidelity}, it is observed that the ratio of missing points can be reduced to an almost negligible degree as long as we configure sufficiently large resolution (\textit{i.e.,} $m$) for PGI generation. In the meantime, however, it is worth reminding that larger image resolution leads to higher representation redundancy, as shown in Table \ref{table:pgi-trade-off}. In practice, we are supposed to flexibly adjust the trade-off between redundancy and accuracy according to specific computational/memory budget and task property.}

\begin{table}[t]
	\renewcommand\arraystretch{1.35}
	\caption{Quantitative neighborhood consistency metrics derived from different choices of $J$ and $\bar{J}$ on ModelNet40, where the number of guidance points (\textit{i.e.,} $N_G$) is uniformly configured as 256.}
	\centering
	\setlength{\tabcolsep}{4.5mm}{
		\begin{tabular}{ c | c | c || c }
			\toprule[1.1pt]
			$N_G$ & $J$ & $\bar{J}$ & Neighborhood Consistency  \\
			\hline
			$256$ & $8$ & $8$ & $47.14$\% \\
			$256$ & $8$ & $16$ & $71.18$\% \\
			$256$ & $8$ & $32$ & $89.31$\% \\
			$256$ & $16$ & $16$ & $49.17$\% \\
			$256$ & $16$ & $32$ & $73.69$\% \\
			$256$ & $16$ & $64$ & $91.66$\% \\
			\bottomrule[1.1pt]
	\end{tabular}}
	\label{table:pgi-neighborhood-consistency}
\end{table}

\textbf{2) Neighborhood Consistency.} \mr{In contrast to the irregularity and unstructuredness of raw point clouds, one major characteristic of the proposed regular geometry representation structure lies in that spatial consistency (\textit{i.e.,} adjacency relations) within local neighborhood should be effectively preserved during the process of 3D-to-2D embedding. More intuitively, neighboring points in the original 3D space are still supposed to be adjacent after being embedded onto the 2D planar space. Here, we need to remind that topological distortions are theoretically inevitable in most cases, unless the target 3D surface is strictly homeomorphic to a 2D plane.}

\mr{We consider a guidance point set $\mathcal{P}_G \in \mathbb{R}^{N_G \times 3}$, which is point-wisely flattened to generate a planar embedding point set $\mathcal{F}_G \in \mathbb{R}^{N_G \times 2}$. Note that there is naturally a (row-wise) one-to-one bijection mapping between $\mathcal{P}_G$ and $\mathcal{F}_G$, \textit{i.e.,} the $i$-th 2D embedding point $\mathbf{f}_{G_i} \in \mathcal{F}_G$ is obtained from the $i$-th 3D guidance point $\mathbf{p}_{G_i} \in \mathcal{P}_G$. Based on this observation, we tend to quantitatively derive the neighborhood consistency metric in the following three steps:}
\begin{enumerate}
\item \mr{we search for $J$ spatial neighbors of $\mathbf{f}_{G_i}$, denoted as $\mathbf{f}_{G_i}^{(j)}$ ($j=1,...,J$), among all 2D embedding points in $\mathcal{F}_G$. Then we can directly locate a 3D guidance point $\mathbf{p}_{G_i}^{(j)}$ that is mapped to $\mathbf{f}_{G_i}^{(j)}$ since they share the same row-wise index in $\mathcal{P}_G$ and $\mathcal{F}_G$, respectively. For convenience, we denote
	$\Omega_i=\left\{\mathbf{p}_{G_i}^{(j)}\right\}_{j=1}^{J}$.}

\item \mr{in the original 3D space, we further search for $\bar{J}$ spatial neighbors of $\mathbf{p}_{G_i}$ among all 3D guidance points in $\mathcal{P}_G$, which are similarly denoted as
	$\bar{\Omega}_i=\left\{\bar{\mathbf{p}}_{G_i}^{(j)}\right\}_{j=1}^{\bar{J}}$.}

\item \mr{we compute the percentage of points in $\Omega_i$ that can be found in $\bar{\Omega}_i$, then deduce the neighborhood consistency metric by iterating the same procedures on all the $N_G$ guidance points (\textit{i.e.,} averaged for $i=1,...,N_G$).}
\end{enumerate}
\mr{It is worth mentioning that local patch parameterizations that operate on context points $\mathcal{C}$ are excluded from the above evaluation process, because in our working mechanism the 2D embeddings of context points are naturally restricted in the scope of the corresponding guidance point, without violating the local adjacency requirement.}

\mr{In the ideal case, where local neighborhood consistency is completely maintained, $\Omega_i$ and $\bar{\Omega}_i$ contain the same set of points (if we set $J = \bar{J}$). In practice, however, when dealing with point clouds with arbitrarily complex geometry and/or topology, there inevitably exist distortions. Hence, we may as well appropriately relax the above evaluation protocol by specifying a larger value for $\bar{J}$, such that $\bar{J}\geq J$.}

\mr{Following such a principle, we experimented with diverse combinations of $J$ and $\bar{J}$ on ModelNet40 (where $N_G=256$). As shown in Table \ref{table:pgi-neighborhood-consistency}, the local neighborhood consistency is effectively preserved during the generation of PGIs.}

\begin{table}[t]
	\renewcommand\arraystretch{1.35}
	\centering
	\caption{Overall accuracy (OA) of different deep shape classification methods on ModelNet40. ``$\ast$'' means that point-wise normals are consumed as additional input attributes.}
	\setlength{\tabcolsep}{6.0mm}{
		\begin{tabular}{ c | c | c }
			\toprule[1.1pt]
			\textbf{Method} & \textbf{\# Points} & \textbf{OA (\%)} \\
			\hline
			PointNet-vanilla \cite{qi2017pointnet} & $1024$ & $87.1$ \\
			PointNet \cite{qi2017pointnet} & $1024$ & $89.2$ \\
			PointNet++ \cite{qi2017pointnet++} $\ast$ & $5000$ & $91.9$ \\
			SpiderCNN \cite{xu2018spidercnn} $\ast$ & $1024$ & $92.4$ \\
			SO-Net \cite{li2018so} $\ast$ & $5000$ & $93.4$ \\
			PointConv \cite{wu2019pointconv} $\ast$ & $1024$ & $92.5$ \\
			DGCNN \cite{wang2019dynamic} & $1024$ & $92.9$ \\
			\hline
			\textbf{\textit{FlatNet-Cls-P}} & $1024$ & $92.6$ \\
			\textbf{\textit{FlatNet-Cls-D}} & $1024$ & $93.4$ \\
			\bottomrule[1.1pt]
	\end{tabular}}
	\label{table:cls-modelnet40}
\end{table}

\begin{table}[t]
	\renewcommand\arraystretch{1.35}
	\centering
	\caption{Ablative analysis of CSConv on ModelNet40 classification.}
	\setlength{\tabcolsep}{5.0mm}{
		\begin{tabular}{ c | c}
			\toprule[1.1pt]
			\textbf{Variant} & \textbf{OA (\%)} \\
			\hline
			\textbf{\textit{FlatNet-Cls-P}} (\textit{w/o} CSConv) & $90.5$ ($-2.1$) \\
			\textbf{\textit{FlatNet-Cls-D}} (\textit{w/o} CSConv) & $92.2$ ($-1.2$) \\
			\hline
			\textbf{\textit{FlatNet-Cls-P}} (\textit{w/} $k \times k$ Conv) & $90.9$ ($-1.7$) \\
			\textbf{\textit{FlatNet-Cls-D}} (\textit{w/} $k \times k$ Conv) & $92.5$ ($-0.9$) \\
			\bottomrule[1.1pt]
	\end{tabular}}
	\label{table:ablation-cls-csconv}
\end{table}

\subsection{Shape Classification} \label{sec:5.2}

\mr{We conducted experiments on the ModelNet40 \cite{wu20153d} dataset, consisting of 12311 synthetic mesh models covering 40 object categories, to benchmark shape classification performances. Under the official split, there are 9843 shapes in the training set and 2468 shapes in the testing set.}

\mr{To quantitatively demonstrate the efficiency and scalability of our method, we actually experimented with an increasing number of input points, \textit{i.e.,} $N=\{1024,2048,5000,10000\}$, uniformly discretized from the original mesh faces. For the creation of PGIs, we configured $N_C=\{12, 24, 50, 100\}$ and $k=\{5, 7, 10, 15\}$ within Flattening-Net to produce different image resolutions, \textit{i.e.,} $m=\{80,112,160,240\}$, respectively. During training, we employed common data augmentation strategies, including random translation, ground-axis rotation, and anisotropic rescaling, to boost the generalization ability. During testing, we did not adopt any voting scheme, which turns to be highly cumbersome and unstable.}

\mr{Table \ref{table:cls-modelnet40} reports the classification accuracy of different point cloud learning frameworks on ModelNet40. Note that our FlatNet-Cls-P and FlatNet-Cls-D are supposed to be regarded as the extensions of the corresponding baseline models PointNet-vanilla \cite{qi2017pointnet} and DGCNN \cite{wang2019dynamic}, respectively, since our task networks are composed of the same building blocks of point-wise MLP \cite{qi2017pointnet} and EdgeConv \cite{wang2019dynamic}. Comparatively, although the PointNet-vanilla baseline with $87.1\%$ accuracy shows limited modeling capacity, our FlatNet-Cls-P variant still achieves much better performance with $92.6\%$ accuracy. For the stronger DGCNN baseline with $92.9\%$ accuracy, our FlatNet-Cls-D variant further brings $0.5\%$ performance gain to reach $93.4\%$ accuracy. Furthermore, we can expect to achieve better performances by replacing more powerful baseline models in addition to \cite{qi2017pointnet,qi2017pointnet,wang2019dynamic}.}

\begin{figure*}[t]
	\centering
	\includegraphics[width=1.0\linewidth]{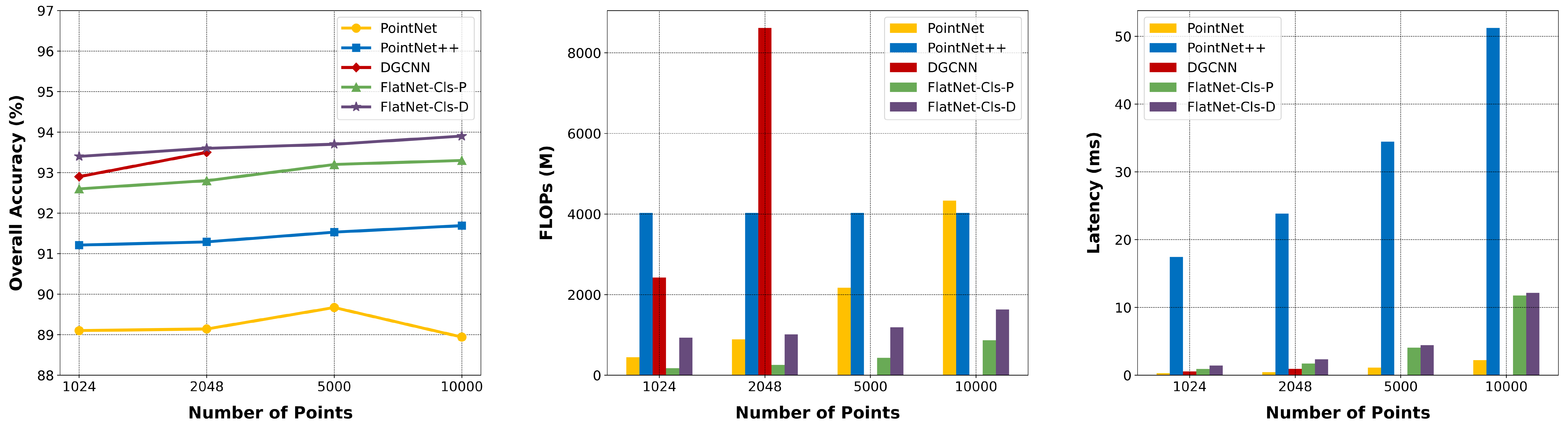}
	\caption{\mr{Comparison of classification accuracy, FLOPs, and latency of different methods when dealing with increasing number of input points.}}
	\label{fig:cls-modelnet40-increasing-points}
\end{figure*}

\mr{Figure \ref{fig:cls-modelnet40-increasing-points} compares the specific statistics of computational efficiency among our learning frameworks and other three representative models \cite{qi2017pointnet,qi2017pointnet++,wang2019dynamic}. For different number of input points, we evaluated accuracy, FLOPs, and latency during the inference stage. Here, we can draw the following several aspects of conclusions:}
\begin{itemize}
	\item \mr{As a representative point-wise convolutional learning paradigm, PointNet \cite{qi2017pointnet} shows satisfactory efficiency for processing sparse point clouds. However, the computational complexity grows linearly as the number of input points increases. More importantly, due to the limited modeling capability and the lack of neighborhood aggregation, consuming denser point clouds as inputs cannot produce higher accuracy, \textit{e.g.,} showing degraded performance when dealing with $10000$ points.}
	\item \mr{Thanks to the hierarchical feature abstraction mechanism, PointNet++ \cite{qi2017pointnet++} achieves stable performance boost for processing denser point clouds. In practice, since its first set abstraction layer uniformly samples $512$ centroids, the measurement of FLOPs is maintained unchanged. However, the downsampling process of FPS can be extremely time-consuming, which limits its scalability to large-scale point clouds.}
	\item \mr{DGCNN \cite{wang2019dynamic} is perhaps the most popular backbone network for point feature extraction, which achieves highly competitive performance and acceptable computational burden for processing sparse point clouds. When we increase the number of input points from $1024$ to $2048$, it can contribute obvious performance gain. However, a major shortcoming is the extremely high computational and memory cost for dense point clouds. Practically, processing a single input model containing $5000$ points requires $5$GB GPU memory in the training phase, making it impossible to converge on ordinary computation devices due to small batch size. Therefore, in Figure \ref{fig:cls-modelnet40-increasing-points}, we did not provide the corresponding statistics for $5000$ and $10000$ points.}
	\item \mr{For both the FlatNet-Cls-P and FlatNet-Cls-D variants, we observed stable performance improvement as well as insignificant growth of FLOPs and latency when dealing with dense inputs.}
\end{itemize}

\begin{table}[t]
	\renewcommand\arraystretch{1.35}
	\centering
	\caption{Performance of real-scanned point cloud object classification on the OBJ\_ONLY and OBJ\_BG settings of ScanObjectNN, in which our Flattening-Net is pretrained on ShapeNetCore and directly applied to generate PGIs on ScanObjectNN without fine-tuning.}
	\setlength{\tabcolsep}{3.0mm}{
		\begin{tabular}{ c | c | c | c }
			\toprule[1.1pt]
			\textbf{Method} & \textbf{\# Points} & \textbf{OBJ\_ONLY} & \textbf{OBJ\_BG} \\
			\hline
			PointNet \cite{qi2017pointnet} & $1024$ & $79.2$ & $73.3$ \\
			PointNet++ \cite{qi2017pointnet++} & $1024$ & $84.3$ & $82.3$ \\
			SpiderCNN \cite{xu2018spidercnn} & $1024$ & $79.5$ & $77.1$ \\
			DGCNN \cite{wang2019dynamic} & $1024$ & $86.2$ & $82.8$ \\
			\hline
			\textbf{\textit{FlatNet-Cls-P}} & $1024$ & $86.5$ & $85.9$ \\
			\textbf{\textit{FlatNet-Cls-D}} & $1024$ & $87.7$ & $86.4$ \\
			\bottomrule[1.1pt]
	\end{tabular}}
	\label{table:transfer-cls-scanobjectnn}
\end{table}

\mr{In addition to evaluating the whole learning pipeline, we performed ablative analysis of the proposed CSConv operator. First, to validate its necessity, we removed the CSConv-driven regional embedding component while maintaining the subsequent task network unchanged. As shown in the first two rows of Table \ref{table:ablation-cls-csconv}, our pipelines suffer from obvious performance degradation. Second, to verify the superiority of our customized surface-style regional embedding procedure over standard convolutional operations, we designed another two variants by replacing CSConv with 2D convolutions with the  kernel size and sliding stride set as $k \times k$. As shown in the last two rows of Table \ref{table:ablation-cls-csconv}, such a modification brings performance boost over the above two baselines to some extent, but is still inferior to the original frameworks.}

\mr{To demonstrate the transferability of Flattening-Net for point cloud parameterization in the learning-based manner, we further performed verification under a transfer learning scenario. Here, we started by pretraining Flattening-Net on ShapeNetCore \cite{chang2015shapenet}, a large-scale synthetic shape repository containing over $50000$ object models. After pretraining, we transferred it to generate PGIs from all point cloud models of the real-scanned ScanObjectNN \cite{uy2019revisiting} dataset without fine-tuning (\textit{i.e.,} the network parameters were fixed), after which we trained the same FlatNet-Cls-P and FlatNet-Cls-D variants for evaluation. Table \ref{table:transfer-cls-scanobjectnn} compares the classification accuracy of different learning frameworks, where our methods show highly competitively performance. Since ScanObjectNN is known to be a much more challenging benchmark dataset, where the object models are typically noisy, incomplete, and accompanied by background points (in the OBJ\_BG setting), the above experiments and comparisons can strongly validate the transferability of Flattening-Net, even for different datasets with big domain gaps.}

\subsection{Semantic Segmentation} \label{sec:5.3}

\mr{We experimented with part segmentation of 3D objects on the ShapeNetPart \cite{yi2016scalable} dataset, which is composed of $16881$ labeled models covering $16$ object categories with totally $50$ different parts. Following the official split, we have $14007$ models for training and $2874$ for testing. In this experiment, each point cloud contains $2048$ points uniformly sampled from the original mesh models. Accordingly, we configured $N_C=24$ and $k=8$ (\textit{i.e.,} $m=128$) within Flattening-Net for the generation of PGIs. As reported in Table \ref{table:part-seg-shapenetpart}, both FlatNet-Seg-P and FlatNet-Seg-D variants outperform their corresponding baselines, \textit{e.g.,} PointNet and DGCNN, with obvious margins.}

\begin{table}[t]
	\renewcommand\arraystretch{1.35}
	\centering
	\caption{Performance of part segmentation on ShapeNetPart measured by mean intersection-over-union (mIoU), where ``$\ast$'' means that point-wise normals are consumed as additional input attributes.}
	\setlength{\tabcolsep}{5.5mm}{
		\begin{tabular}{ c | c | c }
			\toprule[1.1pt]
			\textbf{Method} & \textbf{\# Points} & \textbf{mIoU (\%)} \\
			\hline
			PointNet \cite{qi2017pointnet} & $2048$ & $83.7$ \\
			PointNet++ \cite{qi2017pointnet++} $\ast$ & $2048$ & $85.1$ \\
			SpiderCNN \cite{xu2018spidercnn} $\ast$ & $2048$ & $85.3$ \\
			SO-Net \cite{zhang2019shellnet} $\ast$ & $2048$ & $84.9$ \\
			PointConv \cite{wu2019pointconv} $\ast$ & $2048$ & $85.7$ \\
			DGCNN \cite{wang2019dynamic} & $2048$ & $85.1$ \\
			\hline
			\textbf{\textit{FlatNet-Seg-P}} & $2048$ & $84.9$ \\
			\textbf{\textit{FlatNet-Seg-D}} & $2048$ & $85.8$ \\
			\bottomrule[1.1pt]
	\end{tabular}}
	\label{table:part-seg-shapenetpart}
\end{table}

\begin{table}[t]
	\renewcommand\arraystretch{1.35}
	\centering
	\caption{\mr{Ablative analysis of CSConv on ShapeNetPart segmentation.}}
	\setlength{\tabcolsep}{5.0mm}{
		\mr{\begin{tabular}{ c | c}
			\toprule[1.1pt]
			\textbf{Variant} & \textbf{mIoU (\%)} \\
			\hline
			\textbf{\textit{FlatNet-Seg-P}} (\textit{w/o} CSConv) & $84.1$ ($-0.8$) \\
			\textbf{\textit{FlatNet-Seg-D}} (\textit{w/o} CSConv) & $85.2$ ($-0.6$) \\
			\hline
			\textbf{\textit{FlatNet-Seg-P}} (\textit{w/} $k \times k$ Conv) & $84.5$ ($-0.4$) \\
			\textbf{\textit{FlatNet-Seg-D}} (\textit{w/} $k \times k$ Conv) & $85.5$ ($-0.3$) \\
			\bottomrule[1.1pt]
	\end{tabular}}}
	\label{table:ablation-part-seg-csconv}
\end{table}

\mr{Following the same settings in Section \ref{sec:5.2}, here we also performed ablative analysis of CSConv in the part segmentation scenario. Table \ref{table:ablation-part-seg-csconv} lists the corresponding performances of four different model variants, according to which we can draw consistent conclusions of the necessity and superiority of the regional embedding procedure built upon CSConv.}

\mr{In addition to object-level understanding tasks, we further experimented with scene-level parsing to verify the universality of our PGI-driven point cloud learning paradigm. S3DIS \cite{armeni20163d} is a widely-used indoor scene semantic segmentation dataset for large-scale colored point clouds composed of $271$ single rooms located in $6$ different areas, with over $270$ million densely annotated points covering $13$ semantic classes. Following previous pipelines \cite{hu2021learning}, instead of directly processing raw data, we performed grid sub-sampling and then cropped each complete room into multiple overlapping sub-regions. In the inference phase, we merged predictions on all the cropped sub-regions while performing voting on repeatedly processed points, and then projected the semantic labels of downsampled points to raw data.} \mr{In terms of the generation of PGIs from scene croppings, we followed the same development protocol in the preceding ScanObjectNN classification experiments to pretrain our Flattening-Net on ShapeNetCore and then directly transfer to S3DIS with network parameters fixed.}

\mr{Table \ref{table:indoor-seg-s3dis} reports indoor scene segmentation performance of different methods, among which \cite{landrieu2018large,jiang2019hierarchical,hu2021learning} are particularly specialized for point cloud semantic segmentation scenarios. It is observed that our FlatNet-Scene-Seg variant still achieves competitive performances, which validates the effectiveness of our method when extended to scene data.}

\begin{table}[t]
	\renewcommand\arraystretch{1.35}
	\centering
	\caption{\mr{Performance of large-scale indoor scene segmentation on Area-5 of S3DIS measured by mean class accuracy (mAcc) and mean intersection-over-union (mIoU).}}
	\setlength{\tabcolsep}{5.5mm}{
		\mr{\begin{tabular}{ c | c | c }
			\toprule[1.1pt]
			\textbf{Method} & \textbf{mAcc (\%)} & \textbf{mIoU (\%)} \\
			\hline
			PointNet \cite{qi2017pointnet} & $49.0$ & $41.1$ \\
			PointCNN \cite{li2018pointcnn} & $63.9$ & $57.3$ \\
			SPG \cite{landrieu2018large} & $66.5$ & $58.0$ \\
			HPEIN \cite{jiang2019hierarchical} & $68.3$ & $61.9$ \\
			RandLA-Net \cite{hu2021learning} & $71.5$ & $62.5$ \\
			KPConv \cite{thomas2019kpconv} & $72.8$ & $67.1$ \\
			FPConv \cite{lin2020fpconv} & $68.9$ & $62.8$ \\
			\hline
			\textbf{\textit{FlatNet-Scene-Seg}} & $71.9$ & $62.4$ \\
			\bottomrule[1.1pt]
	\end{tabular}}}
	\label{table:indoor-seg-s3dis}
\end{table}

\subsection{Point Cloud Reconstruction}  \label{sec:5.4}

\mr{We evaluated the learning capacity of different paradigms of deep point auto-encoders in terms of reconstruction quality under the same codeword length. We adopted the same data preparation protocols as introduced in the preceding shape classification experiments on ModelNet40 \cite{wu20153d} to create PGIs from $2048$ and $5000$ points. For comparison, we developed two baseline point cloud auto-encoding frameworks, \textit{i.e.,} an MLP-based \cite{achlioptas2018learning} model called Baseline-Rec-M and a folding-based \cite{yang2018foldingnet} model called Baseline-Rec-F. The former directly regresses point-wise coordinates through a stack of multiple FC layers, while the latter deforms a pre-defined 2D lattice to approximate the target 3D shape.}

\begin{figure}[t]
	\centering
	\includegraphics[width=0.95\linewidth]{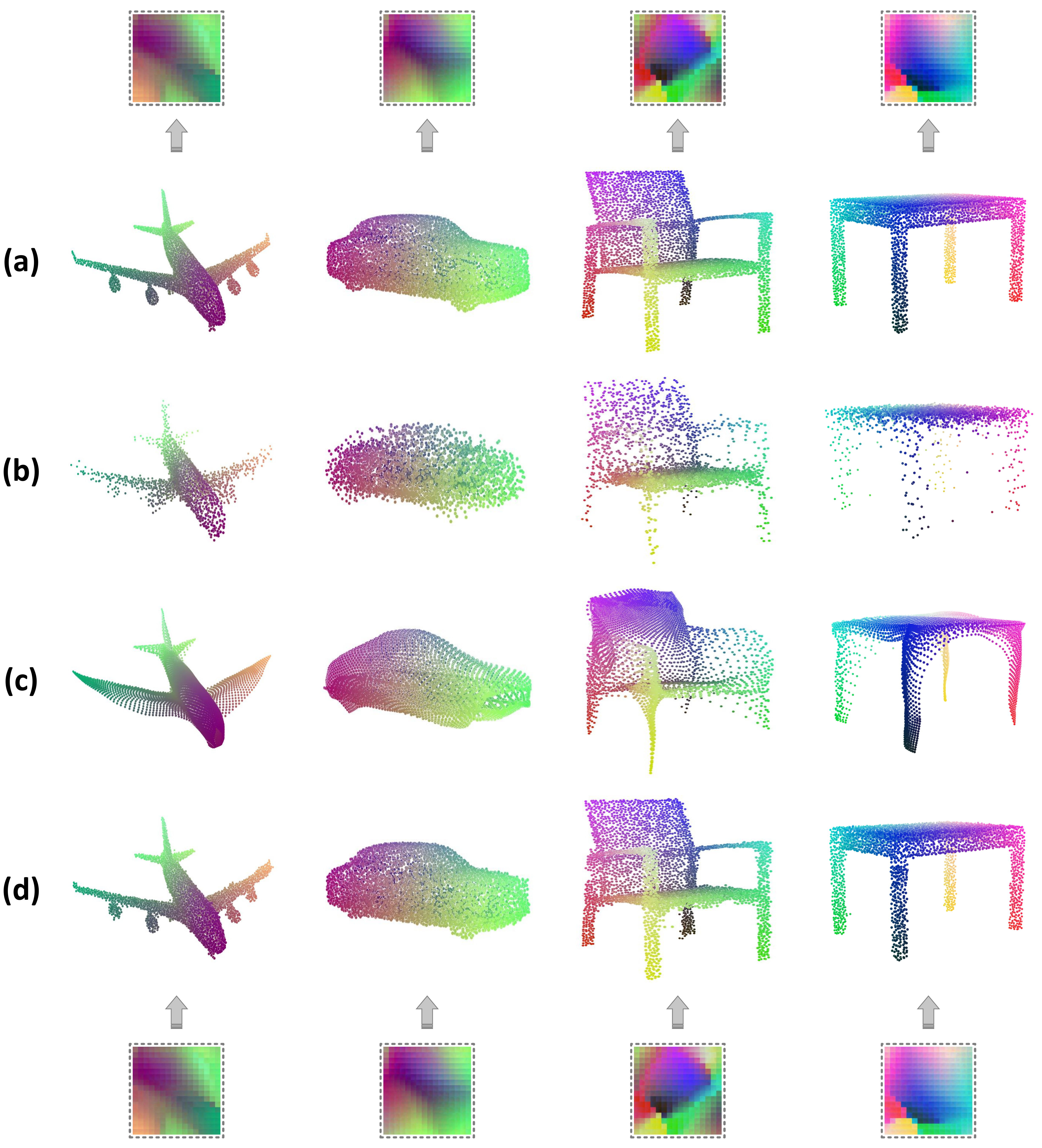}
	\caption{\mr{Visual comparison of point cloud reconstruction results generated by (b) Baseline-Rec-M, (c) Baseline-Rec-F, and (d) our FlatNet-Rec for auto-encoding (a) the input point clouds. In particular, we also show the PGI representation structures corresponding to (a) and (d). }}
	\label{fig:visual-examples-rec}
\end{figure}

\mr{Table \ref{table:rec-modelnet40} quantitatively compares the reconstruction quality as well as model sizes of different methods. For MLP-based frameworks, the corresponding network complexity grows significantly when dealing with dense point clouds because the number of output neurons exactly relies on the required number of reconstructed points, resulting in greater learning difficulty. For folding-based frameworks, the same network configuration can be applied to reconstruct different number of points while achieving better reconstruction quality under both sparse and dense input settings. Comparatively, our FlatNet-Rec variant produces the lowest reconstruction errors with moderate model size. Figure \ref{fig:visual-examples-rec} visually compares the reconstruction results obtained by different methods, in which it can be observed that our results are closer to input shapes with less noises and outliers.}

\begin{table}[t]
	\renewcommand\arraystretch{1.35}
	\centering
	\caption{\mr{Quantitative performance of point cloud reconstruction on ModelNet40 measured by Chamfer distance (CD) and model size (MS) corresponding to different number of input points.}}
	\setlength{\tabcolsep}{3.5mm}{
		\mr{\begin{tabular}{  c | c | c | c }
			\toprule[1.1pt]
			\textbf{Method} & \textbf{\# Points}  & \textbf{CD ($\mathbf{10^{-3}}$)} & \textbf{MS (MB)} \\
			\hline
			Baseline-Rec-M & $2048$ & $1.75$ & $60$ \\
			Baseline-Rec-M & $5000$ & $1.61$ & $322$ \\
			\hline
			Baseline-Rec-F & $2048$ & $1.69$ & $37$ \\
			Baseline-Rec-F & $5000$ & $1.22$ & $37$ \\
			\hline
			\textbf{\textit{FlatNet-Rec}} & $2048$ & $0.93$ & $35$ \\
			\textbf{\textit{FlatNet-Rec}} & $5000$ & $0.85$ & $35$ \\
			\bottomrule[1.1pt]
	\end{tabular}}}
	\label{table:rec-modelnet40}
\end{table}

\subsection{Point Cloud Upsampling}  \label{sec:5.5}

\begin{figure}[t]
	\centering
	\includegraphics[width=0.90\linewidth]{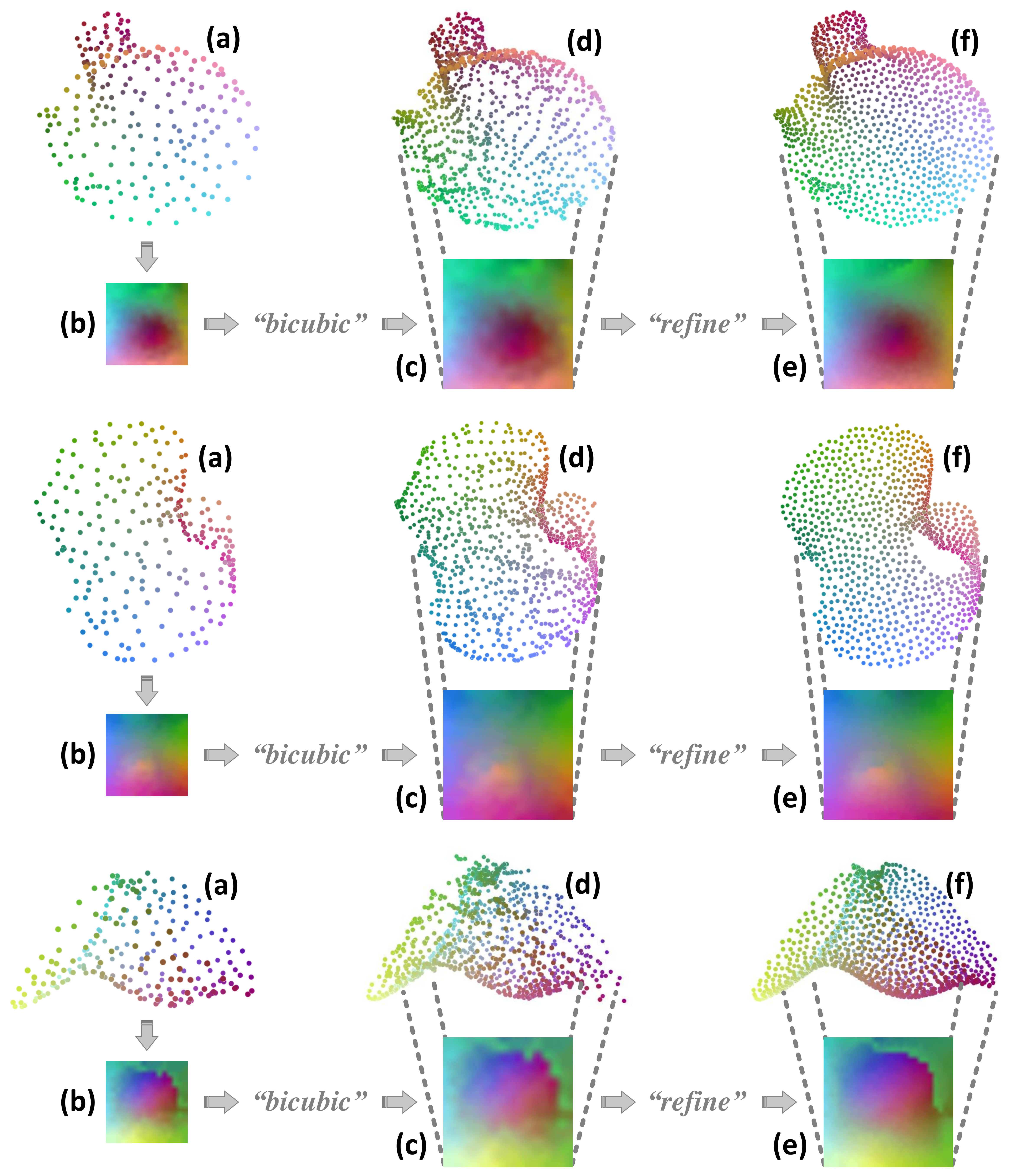}
	\caption{\mr{Illustration of the processing pipeline of FlatNet-Ups that implements 3D point cloud upsampling as 2D PGI super-resolution. Given (a) an input sparse point cloud patch, we generate (b) an LR-PGI and then apply standard bicubic image interpolation to obtain (c) an enlarged LR-PGI, which corresponds to (d) the coarsely upsampled patch. After that, the initial enlarged LR-PGI is refined into the resulting (e) HR-PGI, which corresponds to the desired (f) dense upsampling result.}}
	\label{fig:visual-examples-patch-pgi-sr}
\end{figure}

\mr{We conducted experiments on $4 \times$ point cloud upsampling using the same dataset as \cite{li2019pu} (which we call PU147), where there are $147$ training models and $27$ testing models. Under the same development protocol, the numbers of points in the input sparse model and the target ground-truth model are $2048$ and $8192$, respectively. In the actual training and inference stages, input patches uniformly contain $256$ points. Figure \ref{fig:visual-examples-patch-pgi-sr} illustrates the workflow of upsampling sparse local patches through PGI super-resolution.}

\begin{figure*}[t]
	\centering
	\includegraphics[width=0.90\linewidth]{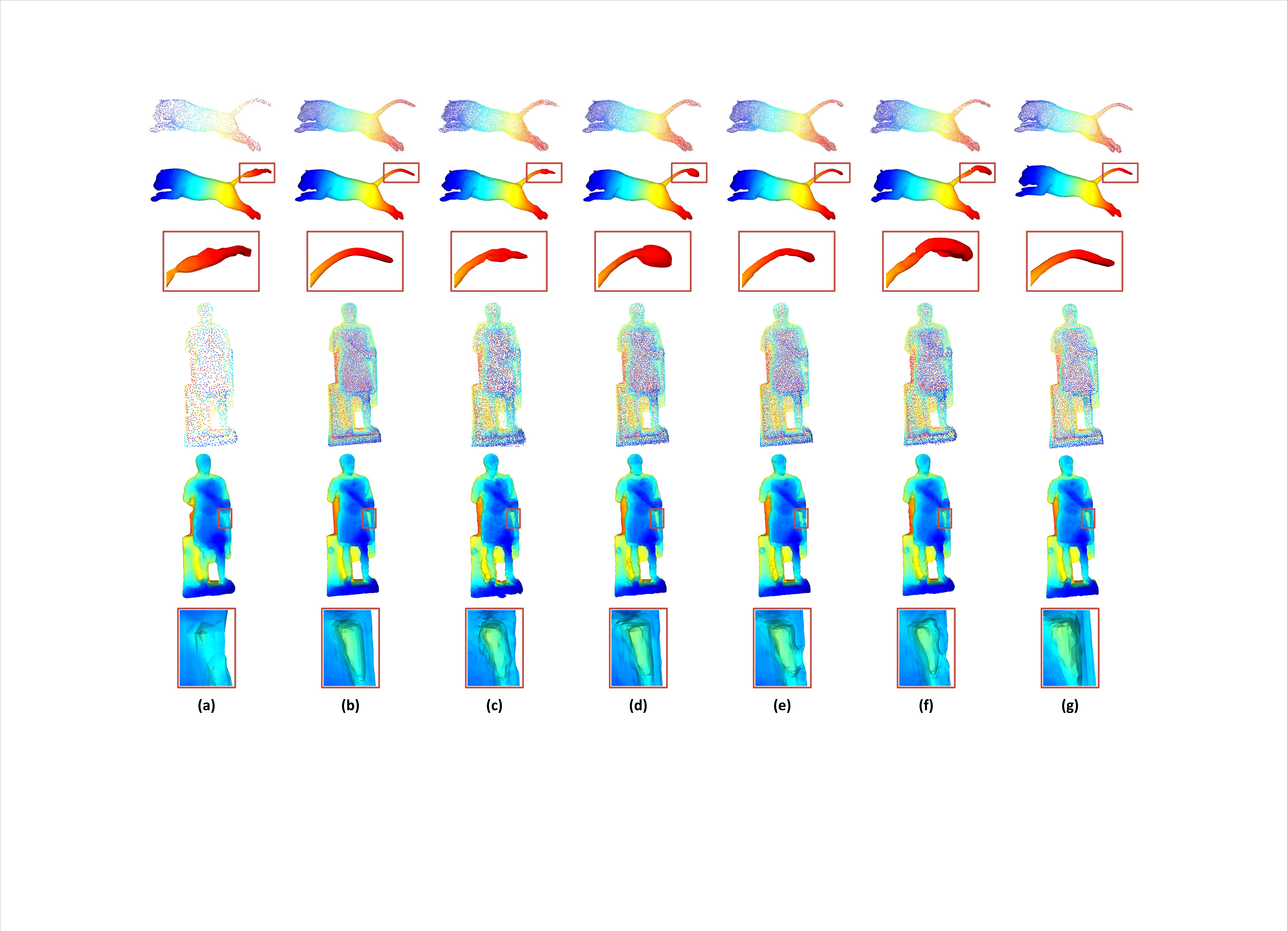}
	\caption{\mr{Visual comparison of point cloud upsampling results. Given (a) input sparse point clouds and (b) dense ground-truths, we present typical upsampling examples generated by (c) PU-Net \cite{yu2018pu}, (d) MPU \cite{yifan2019patch}, (e) PU-GAN \cite{li2019pu}, (f) PUGeo-Net \cite{qian2020pugeo}, and (g) our FlatNet-Ups.}}
	\label{fig:visual-examples-ups}
\end{figure*}

\mr{Table \ref{table:ups-pu147} compares our upsampling network of FlatNet-Ups with previous deep learning-based frameworks that are specialized for point cloud upsampling. It can be observed that our PGI-driven framework outperforms the other methods in terms of all the three evaluation metrics with obvious margins. Figure \ref{fig:visual-examples-ups} provides some visual comparisons of the upsampling results obtained by different methods, in which our results show better surface mesh reconstruction quality. In fact, in our implementation, we only incorporated some fundamental design experience from the research community of image super-resolution \cite{wang2020deep}. Still, our experimental results have already demonstrated the potential of adapting 2D image processing techniques for our PGI representations. We reasonably expect that further performance improvement can be achieved by introducing more advanced and specialized image-domain learning modules.}

\begin{table}
	\renewcommand\arraystretch{1.35}
	\centering
	\caption{\mr{Quantitative performance of point cloud upsampling on PU147 measured by Chamfer distance (CD), Hausdorff distance (HD), and point-to-surface (P2F) distance.}}
	\setlength{\tabcolsep}{2.75mm}{
		\begin{tabular}{ c | c | c | c}
			\toprule[1.1pt]
			\textbf{Method} & \textbf{P2F ($\mathbf{10^{-3}}$)} & \textbf{CD ($\mathbf{10^{-3}}$)} & \textbf{HD ($\mathbf{10^{-3}}$)} \\
			\hline
			PU-Net \cite{yu2018pu} & $6.97$ & $0.72$ & $8.93$ \\
			MPU \cite{yifan2019patch} & $3.93$ & $0.49$ & $6.06$ \\
			PU-GAN \cite{li2019pu} & $2.40$ & $0.29$ & $4.75$ \\
			PUGeo-Net \cite{qian2020pugeo} & $2.85$ & $0.32$ & $3.28$ \\
			\hline
			\mr{\textbf{\textit{FlatNet-Ups}}} & \mr{$2.11$} & \mr{$0.25$} & \mr{$2.93$} \\
			\bottomrule[1.1pt]
	\end{tabular}}
	\label{table:ups-pu147}
\end{table}

\begin{table}[t]
	\renewcommand\arraystretch{1.35}
	\caption{Comparison of regular 2D representation-based 3D shape recognition frameworks on ModelNet40 classification with different input types of classic GIs, spherical parameterizations (S.P.) produced by equirectangular projection, and the proposed PGIs.}
	\begin{center}
		\setlength{\tabcolsep}{4.75mm}{
			\begin{tabular}{ c | c | c }
				\toprule[1.1pt]
				\textbf{Method} & \textbf{Para. Type} & \textbf{OA (\%)} \\
				\hline
				DLGI \cite{sinha2016deep}  & GIs & $83.9$ \\
				SNGC \cite{haim2019surface} & GIs & $91.6$ \\
				\hline
				EP-Cls-P & S.P. & $88.5$ \\
				EP-Cls-I & S.P. & $90.1$ \\
				\hline
				\textbf{\textit{FlatNet-Cls-P}} & PGIs & $92.6$ \\
				\textbf{\textit{FlatNet-Cls-D}} & PGIs & $93.4$ \\
				\bottomrule[1.1pt]
		\end{tabular}}
	\end{center}
	\label{table:para-cls-modelnet40}
\end{table}

\section{Differences between PGIs and GIs} \label{sec:diff-pgi-gi}

\mr{Essentially, both the proposed point geometry image (PGI) and the traditional geometry image (GI) \cite{gu2002geometry} are designed for regular 2D representation of 3D geometric information, which produce a three-channel colored image at the output end. Nevertheless, we emphasize that PGIs are fundamentally different from GIs in terms of generation and target domain. More specifically, GIs are created from meshes and generated by optimization-based surface parameterization algorithms \cite{DBLP:journals/tog/BommesZK09, DBLP:journals/cgf/LiuZXGG08, DBLP:journals/tog/SpringbornSP08, DBLP:journals/tog/ShefferLMB05, DBLP:journals/cgf/ZhaoSLZYLWGG20, DBLP:journals/tvcg/JinKLG08}, which are able to compute high-quality GIs from 3D shapes with relatively simple geometry and topology. In practice, the generation pipeline of GIs is often used for GPU-accelerated rendering and texture mapping in applications of movies and video games. Increasing the complexity of geometry/topology can pose significant challenges to these methods. Moreover, the above mesh-oriented computational process is restricted to manifold meshes, despite the fact that most real-world 3D models are non-manifolds. These challenges can seriously diminish the application of classic GIs in 3D geometric deep learning. By contrast, our method is particularly developed for unstructured point clouds with arbitrary geometry and topology, working for both manifolds and non-manifolds. As a generic representation modality for 3D deep learning, our PGIs support a much wider range of downstream point cloud processing and understanding scenarios.}

\mr{As mentioned in Section \ref{sec-2-1}, previous studies \cite{sinha2016deep, haim2019surface} also explore similar deep learning-based shape recognition pipelines where off-the-shelf 2D CNNs are directly applied to classic GIs, as shown in the first two rows of Table \ref{table:para-cls-modelnet40}. In addition, we further experimented with two equirectangular projection (EP)-driven baseline frameworks, as shown in the middle two rows of Table \ref{table:para-cls-modelnet40}. The variant of EP-Cls-P is modified from our preceding FlatNet-Cls-P by changing the input signals from PGIs to spherical parameterizations produced by EP while maintaining all the other components. For the variant of EP-Cls-I, as conducted in \cite{haim2019surface}, we employ Inception-V3 \cite{szegedy2016rethinking} for image feature extraction. We can draw some useful conclusions from the above comparisons. First, the two EP-based baselines underperform our methods, in that local neighborhood consistency and manifold property are greatly destroyed, which hinders effective feature aggregation. Second, classic GIs turn to be sub-optimal when used as inputs for the subsequent deep networks. We reason that this is caused by the loss of geometric structure during the conversion from raw data to genus-zero manifold meshes.}

\section{Conclusion} \label{sec:conclusion}

\mr{This paper focuses on regular 2D representation for irregular 3D geometry of unstructured point clouds. We proposed an unsupervised learning architecture, namely Flattening-Net, to convert arbitrary point clouds into PGI structures, capturing spatial coordinates in image pixels while effectively preserving local neighborhood consistency. Accordingly, we further developed CSConv, a novel surface-style point convolution operator, which achieves efficient and scalable regional embedding. We demonstrated the effectiveness of Flattening-Net by applying PGIs to diverse point cloud processing and understanding applications, in which our frameworks show highly competitive performance, although our major goal is not to pursue state-of-the-arts in all involved application scenarios by designing various fancy and complicated task-specific network architectures.}

\mr{In conclusion, our extensive experimental results have convincingly indicated the potential and universality of PGIs in 3D deep learning. We believe that such a regular geometry representation modality will open up many new possibilities in the point cloud community. In the future, we plan to extend the scope of our geometry parameterization approach from static 3D point clouds to dynamic sequences while preserving spatio-temporal correspondence between consecutive frames. In terms of downstream applications, it is promising to construct PGI-based point cloud compression frameworks that are highly desired in various practical scenarios \cite{schwarz2018emerging}, where mature 2D image/video codecs can be seamlessly introduced, as done in \cite{hou2014highly,hou2014compressing}.}


\begin{thebibliography}{10}
\providecommand{\url}[1]{#1}
\csname url@samestyle\endcsname
\providecommand{\newblock}{\relax}
\providecommand{\bibinfo}[2]{#2}
\providecommand{\BIBentrySTDinterwordspacing}{\spaceskip=0pt\relax}
\providecommand{\BIBentryALTinterwordstretchfactor}{4}
\providecommand{\BIBentryALTinterwordspacing}{\spaceskip=\fontdimen2\font plus
\BIBentryALTinterwordstretchfactor\fontdimen3\font minus
  \fontdimen4\font\relax}
\providecommand{\BIBforeignlanguage}[2]{{%
\expandafter\ifx\csname l@#1\endcsname\relax
\typeout{** WARNING: IEEEtran.bst: No hyphenation pattern has been}%
\typeout{** loaded for the language `#1'. Using the pattern for}%
\typeout{** the default language instead.}%
\else
\language=\csname l@#1\endcsname
\fi
#2}}
\providecommand{\BIBdecl}{\relax}
\BIBdecl

\bibitem{krizhevsky2012imagenet}
A.~Krizhevsky, I.~Sutskever, and G.~E. Hinton, ``Imagenet classification with
  deep convolutional neural networks,'' \emph{Proc. NeurIPS}, vol.~25, pp.
  1097--1105, 2012.

\bibitem{karpathy2014large}
A.~Karpathy, G.~Toderici, S.~Shetty, T.~Leung, R.~Sukthankar, and L.~Fei-Fei,
  ``Large-scale video classification with convolutional neural networks,'' in
  \emph{Proc. CVPR}, 2014, pp. 1725--1732.

\bibitem{tran2015learning}
D.~Tran, L.~Bourdev, R.~Fergus, L.~Torresani, and M.~Paluri, ``Learning
  spatiotemporal features with 3d convolutional networks,'' in \emph{Proc.
  ICCV}, 2015, pp. 4489--4497.

\bibitem{long2015fully}
J.~Long, E.~Shelhamer, and T.~Darrell, ``Fully convolutional networks for
  semantic segmentation,'' in \emph{Proc. CVPR}, 2015, pp. 3431--3440.

\bibitem{he2016deep}
K.~He, X.~Zhang, S.~Ren, and J.~Sun, ``Deep residual learning for image
  recognition,'' in \emph{Proc. CVPR}, 2016, pp. 770--778.

\bibitem{simonyan2014very}
K.~Simonyan and A.~Zisserman, ``Very deep convolutional networks for
  large-scale image recognition,'' \emph{Proc. ICLR}, 2015.

\bibitem{szegedy2015going}
C.~Szegedy, W.~Liu, Y.~Jia, P.~Sermanet, S.~Reed, D.~Anguelov, D.~Erhan,
  V.~Vanhoucke, and A.~Rabinovich, ``Going deeper with convolutions,'' in
  \emph{Proc. CVPR}, 2015, pp. 1--9.

\bibitem{szegedy2016rethinking}
C.~Szegedy, V.~Vanhoucke, S.~Ioffe, J.~Shlens, and Z.~Wojna, ``Rethinking the
  inception architecture for computer vision,'' in \emph{Proc. CVPR}, 2016, pp.
  2818--2826.

\bibitem{xie2017aggregated}
S.~Xie, R.~Girshick, P.~Doll{\'a}r, Z.~Tu, and K.~He, ``Aggregated residual
  transformations for deep neural networks,'' in \emph{Proc. CVPR}, 2017, pp.
  1492--1500.

\bibitem{huang2017densely}
G.~Huang, Z.~Liu, L.~Van Der~Maaten, and K.~Q. Weinberger, ``Densely connected
  convolutional networks,'' in \emph{Proc. CVPR}, 2017, pp. 4700--4708.

\bibitem{tan2019efficientnet}
M.~Tan and Q.~Le, ``Efficientnet: Rethinking model scaling for convolutional
  neural networks,'' in \emph{Proc. ICML}, 2019, pp. 6105--6114.

\bibitem{maturana2015voxnet}
D.~Maturana and S.~Scherer, ``Voxnet: A 3d convolutional neural network for
  real-time object recognition,'' in \emph{Proc. IROS}, 2015, pp. 922--928.

\bibitem{qi2016volumetric}
C.~R. Qi, H.~Su, M.~Nie{\ss}ner, A.~Dai, M.~Yan, and L.~J. Guibas, ``Volumetric
  and multi-view cnns for object classification on 3d data,'' in \emph{Proc.
  CVPR}, 2016, pp. 5648--5656.

\bibitem{wu20153d}
Z.~Wu, S.~Song, A.~Khosla, F.~Yu, L.~Zhang, X.~Tang, and J.~Xiao, ``3d
  shapenets: A deep representation for volumetric shapes,'' in \emph{Proc.
  CVPR}, 2015, pp. 1912--1920.

\bibitem{riegler2017octnet}
G.~Riegler, A.~Osman~Ulusoy, and A.~Geiger, ``Octnet: Learning deep 3d
  representations at high resolutions,'' in \emph{Proc. CVPR}, 2017, pp.
  3577--3586.

\bibitem{wang2017cnn}
P.-S. Wang, Y.~Liu, Y.-X. Guo, C.-Y. Sun, and X.~Tong, ``O-cnn: Octree-based
  convolutional neural networks for 3d shape analysis,'' \emph{ACM Trans.
  Graph.}, vol.~36, no.~4, pp. 1--11, 2017.

\bibitem{klokov2017escape}
R.~Klokov and V.~Lempitsky, ``Escape from cells: Deep kd-networks for the
  recognition of 3d point cloud models,'' in \emph{Proc. ICCV}, 2017, pp.
  863--872.

\bibitem{su2015multi}
H.~Su, S.~Maji, E.~Kalogerakis, and E.~Learned-Miller, ``Multi-view
  convolutional neural networks for 3d shape recognition,'' in \emph{Proc.
  CVPR}, 2015, pp. 945--953.

\bibitem{kalogerakis20173d}
E.~Kalogerakis, M.~Averkiou, S.~Maji, and S.~Chaudhuri, ``3d shape segmentation
  with projective convolutional networks,'' in \emph{Proc. CVPR}, 2017, pp.
  3779--3788.

\bibitem{yu2018multi}
T.~Yu, J.~Meng, and J.~Yuan, ``Multi-view harmonized bilinear network for 3d
  object recognition,'' in \emph{Proc. CVPR}, 2018, pp. 186--194.

\bibitem{kanezaki2018rotationnet}
A.~Kanezaki, Y.~Matsushita, and Y.~Nishida, ``Rotationnet: Joint object
  categorization and pose estimation using multiviews from unsupervised
  viewpoints,'' in \emph{Proc. CVPR}, 2018, pp. 5010--5019.

\bibitem{qi2017pointnet}
C.~R. Qi, H.~Su, K.~Mo, and L.~J. Guibas, ``Pointnet: Deep learning on point
  sets for 3d classification and segmentation,'' in \emph{Proc. CVPR}, 2017,
  pp. 652--660.

\bibitem{qi2017pointnet++}
C.~R. Qi, L.~Yi, H.~Su, and L.~J. Guibas, ``Pointnet++: Deep hierarchical
  feature learning on point sets in a metric space,'' in \emph{Proc. NeurIPS},
  2017, pp. 5105--5114.

\bibitem{groh2018flex}
F.~Groh, P.~Wieschollek, and H.~Lensch, ``Flex-convolution,'' in \emph{Proc.
  ACCV}, 2018, pp. 105--122.

\bibitem{hua2018pointwise}
B.-S. Hua, M.-K. Tran, and S.-K. Yeung, ``Pointwise convolutional neural
  networks,'' in \emph{Proc. CVPR}, 2018, pp. 984--993.

\bibitem{li2018so}
J.~Li, B.~M. Chen, and G.~H. Lee, ``So-net: Self-organizing network for point
  cloud analysis,'' in \emph{Proc. CVPR}, 2018, pp. 9397--9406.

\bibitem{su2018splatnet}
H.~Su, V.~Jampani, D.~Sun, S.~Maji, E.~Kalogerakis, M.-H. Yang, and J.~Kautz,
  ``Splatnet: Sparse lattice networks for point cloud processing,'' in
  \emph{Proc. CVPR}, 2018, pp. 2530--2539.

\bibitem{xu2018spidercnn}
Y.~Xu, T.~Fan, M.~Xu, L.~Zeng, and Y.~Qiao, ``Spidercnn: Deep learning on point
  sets with parameterized convolutional filters,'' in \emph{Proc. ECCV}, 2018,
  pp. 87--102.

\bibitem{li2018pointcnn}
Y.~Li, R.~Bu, M.~Sun, W.~Wu, X.~Di, and B.~Chen, ``Pointcnn: Convolution on
  $\chi$-transformed points,'' in \emph{Proc. NeurIPS}, 2018, pp. 828--838.

\bibitem{wu2019pointconv}
W.~Wu, Z.~Qi, and L.~Fuxin, ``Pointconv: Deep convolutional networks on 3d
  point clouds,'' in \emph{Proc. CVPR}, 2019, pp. 9621--9630.

\bibitem{liu2019relation}
Y.~Liu, B.~Fan, S.~Xiang, and C.~Pan, ``Relation-shape convolutional neural
  network for point cloud analysis,'' in \emph{Proc. CVPR}, 2019, pp.
  8895--8904.

\bibitem{thomas2019kpconv}
H.~Thomas, C.~R. Qi, J.-E. Deschaud, B.~Marcotegui, F.~Goulette, and L.~J.
  Guibas, ``Kpconv: Flexible and deformable convolution for point clouds,'' in
  \emph{Proc. ICCV}, 2019, pp. 6411--6420.

\bibitem{zhang2019shellnet}
Z.~Zhang, B.-S. Hua, and S.-K. Yeung, ``Shellnet: Efficient point cloud
  convolutional neural networks using concentric shells statistics,'' in
  \emph{Proc. ICCV}, 2019, pp. 1607--1616.

\bibitem{verma2018feastnet}
N.~Verma, E.~Boyer, and J.~Verbeek, ``Feastnet: Feature-steered graph
  convolutions for 3d shape analysis,'' in \emph{Proc. CVPR}, 2018, pp.
  2598--2606.

\bibitem{wang2019dynamic}
Y.~Wang, Y.~Sun, Z.~Liu, S.~E. Sarma, M.~M. Bronstein, and J.~M. Solomon,
  ``Dynamic graph cnn for learning on point clouds,'' \emph{ACM Trans. Graph.},
  vol.~38, no.~5, pp. 1--12, 2019.

\bibitem{liu2019pvcnn}
Z.~Liu, H.~Tang, Y.~Lin, and S.~Han, ``Point-voxel cnn for efficient 3d deep
  learning,'' in \emph{Proc. NeurIPS}, 2019.

\bibitem{xu2020grid}
Q.~Xu, X.~Sun, C.-Y. Wu, P.~Wang, and U.~Neumann, ``Grid-gcn for fast and
  scalable point cloud learning,'' in \emph{Proc. CVPR}, 2020, pp. 5661--5670.

\bibitem{masci2015geodesic}
J.~Masci, D.~Boscaini, M.~Bronstein, and P.~Vandergheynst, ``Geodesic
  convolutional neural networks on riemannian manifolds,'' in \emph{Proc. ICCV
  Workshop}, 2015, pp. 37--45.

\bibitem{boscaini2016learning}
D.~Boscaini, J.~Masci, E.~Rodoi{\`a}, and M.~Bronstein, ``Learning shape
  correspondence with anisotropic convolutional neural networks,'' in
  \emph{Proc. NeurIPS}, 2016, pp. 3197--3205.

\bibitem{monti2017geometric}
F.~Monti, D.~Boscaini, J.~Masci, E.~Rodola, J.~Svoboda, and M.~M. Bronstein,
  ``Geometric deep learning on graphs and manifolds using mixture model cnns,''
  in \emph{Proc. CVPR}, 2017, pp. 5115--5124.

\bibitem{sinha2016deep}
A.~Sinha, J.~Bai, and K.~Ramani, ``Deep learning 3d shape surfaces using
  geometry images,'' in \emph{Proc. ECCV}, 2016, pp. 223--240.

\bibitem{gu2002geometry}
X.~Gu, S.~J. Gortler, and H.~Hoppe, ``Geometry images,'' \emph{Proc. SIGGRAPH},
  vol.~21, no.~3, pp. 355--361, 2002.

\bibitem{sinha2017surfnet}
A.~Sinha, A.~Unmesh, Q.~Huang, and K.~Ramani, ``Surfnet: Generating 3d shape
  surfaces using deep residual networks,'' in \emph{Proc. CVPR}, 2017, pp.
  6040--6049.

\bibitem{maron2017convolutional}
H.~Maron, M.~Galun, N.~Aigerman, M.~Trope, N.~Dym, E.~Yumer, V.~G. Kim, and
  Y.~Lipman, ``Convolutional neural networks on surfaces via seamless toric
  covers,'' \emph{ACM Trans. Graph.}, vol.~36, no.~4, pp. 71--1, 2017.

\bibitem{haim2019surface}
N.~Haim, N.~Segol, H.~Ben-Hamu, H.~Maron, and Y.~Lipman, ``Surface networks via
  general covers,'' in \emph{Proc. ICCV}, 2019, pp. 632--641.

\bibitem{ezuz2017gwcnn}
D.~Ezuz, J.~Solomon, V.~G. Kim, and M.~Ben-Chen, ``Gwcnn: A metric alignment
  layer for deep shape analysis,'' \emph{Comput. Graph. Forum}, vol.~36, no.~5,
  pp. 49--57, 2017.

\bibitem{tatarchenko2018tangent}
M.~Tatarchenko, J.~Park, V.~Koltun, and Q.-Y. Zhou, ``Tangent convolutions for
  dense prediction in 3d,'' in \emph{Proc. CVPR}, 2018, pp. 3887--3896.

\bibitem{lin2020fpconv}
Y.~Lin, Z.~Yan, H.~Huang, D.~Du, L.~Liu, S.~Cui, and X.~Han, ``Fpconv: Learning
  local flattening for point convolution,'' in \emph{Proc. CVPR}, 2020, pp.
  4293--4302.

\bibitem{komarichev2019cnn}
A.~Komarichev, Z.~Zhong, and J.~Hua, ``A-cnn: Annularly convolutional neural
  networks on point clouds,'' in \emph{Proc. CVPR}, 2019, pp. 7421--7430.

\bibitem{cao20173d}
Z.~Cao, Q.~Huang, and R.~Karthik, ``3d object classification via spherical
  projections,'' in \emph{3DV}, 2017, pp. 566--574.

\bibitem{esteves2018learning}
C.~Esteves, C.~Allen-Blanchette, A.~Makadia, and K.~Daniilidis, ``Learning so
  (3) equivariant representations with spherical cnns,'' in \emph{Proc. ECCV},
  2018, pp. 52--68.

\bibitem{coors2018spherenet}
B.~Coors, A.~P. Condurache, and A.~Geiger, ``Spherenet: Learning spherical
  representations for detection and classification in omnidirectional images,''
  in \emph{Proc. ECCV}, 2018, pp. 518--533.

\bibitem{cohen2018spherical}
T.~S. Cohen, M.~Geiger, J.~Köhler, and M.~Welling, ``Spherical {CNN}s,'' in
  \emph{Proc. ICLR}, 2018.

\bibitem{kondor2018clebsch}
R.~Kondor, Z.~Lin, and S.~Trivedi, ``Clebsch--gordan nets: a fully fourier
  space spherical convolutional neural network,'' \emph{Proc. NeurIPS},
  vol.~31, 2018.

\bibitem{jiang2018spherical}
C.~M. Jiang, J.~Huang, K.~Kashinath, Prabhat, P.~Marcus, and M.~Niessner,
  ``Spherical {CNN}s on unstructured grids,'' in \emph{Proc. ICLR}, 2019.

\bibitem{cohen2019gauge}
T.~Cohen, M.~Weiler, B.~Kicanaoglu, and M.~Welling, ``Gauge equivariant
  convolutional networks and the icosahedral cnn,'' in \emph{Proc. ICML}, 2019,
  pp. 1321--1330.

\bibitem{rao2019spherical}
Y.~Rao, J.~Lu, and J.~Zhou, ``Spherical fractal convolutional neural networks
  for point cloud recognition,'' in \emph{Proc. CVPR}, 2019, pp. 452--460.

\bibitem{yang2018foldingnet}
Y.~Yang, C.~Feng, Y.~Shen, and D.~Tian, ``Foldingnet: Point cloud auto-encoder
  via deep grid deformation,'' in \emph{Proc. CVPR}, 2018, pp. 206--215.

\bibitem{groueix2018papier}
T.~Groueix, M.~Fisher, V.~G. Kim, B.~C. Russell, and M.~Aubry, ``A
  papier-m{\^a}ch{\'e} approach to learning 3d surface generation,'' in
  \emph{Proc. CVPR}, 2018, pp. 216--224.

\bibitem{chen2019deep}
S.~Chen, C.~Duan, Y.~Yang, D.~Li, C.~Feng, and D.~Tian, ``Deep unsupervised
  learning of 3d point clouds via graph topology inference and filtering,''
  \emph{IEEE Trans. Image Process.}, vol.~29, pp. 3183--3198, 2019.

\bibitem{deprelle2019learning}
T.~Deprelle, T.~Groueix, M.~Fisher, V.~G. Kim, B.~C. Russell, and M.~Aubry,
  ``Learning elementary structures for 3d shape generation and matching,'' in
  \emph{Proc. NeurIPS}, 2019, pp. 7433--7443.

\bibitem{pang2021tearingnet}
J.~Pang, D.~Li, and D.~Tian, ``Tearingnet: Point cloud autoencoder to learn
  topology-friendly representations,'' in \emph{Proc. CVPR}, 2021, pp.
  7453--7462.

\bibitem{yi2016scalable}
L.~Yi, V.~G. Kim, D.~Ceylan, I.-C. Shen, M.~Yan, H.~Su, C.~Lu, Q.~Huang,
  A.~Sheffer, and L.~Guibas, ``A scalable active framework for region
  annotation in 3d shape collections,'' \emph{ACM Trans. Graph.}, vol.~35,
  no.~6, pp. 1--12, 2016.

\bibitem{guo2020deep}
Y.~Guo, H.~Wang, Q.~Hu, H.~Liu, L.~Liu, and M.~Bennamoun, ``Deep learning for
  3d point clouds: A survey,'' \emph{IEEE Trans. Pattern Anal. Mach. Intell.},
  2020.

\bibitem{wu2016learning}
J.~Wu, C.~Zhang, T.~Xue, W.~T. Freeman, and J.~B. Tenenbaum, ``Learning a
  probabilistic latent space of object shapes via 3d generative-adversarial
  modeling,'' in \emph{Proc. NeurIPS}, 2016, pp. 82--90.

\bibitem{li2017grass}
J.~Li, K.~Xu, S.~Chaudhuri, E.~Yumer, H.~Zhang, and L.~Guibas, ``Grass:
  Generative recursive autoencoders for shape structures,'' \emph{ACM Trans.
  Graph.}, vol.~36, no.~4, pp. 1--14, 2017.

\bibitem{achlioptas2018learning}
P.~Achlioptas, O.~Diamanti, I.~Mitliagkas, and L.~Guibas, ``Learning
  representations and generative models for 3d point clouds,'' in \emph{Proc.
  ICML}, 2018, pp. 40--49.

\bibitem{park2019deepsdf}
J.~J. Park, P.~Florence, J.~Straub, R.~Newcombe, and S.~Lovegrove, ``Deepsdf:
  Learning continuous signed distance functions for shape representation,'' in
  \emph{Proc. CVPR}, 2019, pp. 165--174.

\bibitem{mescheder2019occupancy}
L.~Mescheder, M.~Oechsle, M.~Niemeyer, S.~Nowozin, and A.~Geiger, ``Occupancy
  networks: Learning 3d reconstruction in function space,'' in \emph{Proc.
  CVPR}, 2019, pp. 4460--4470.

\bibitem{chen2019learning}
Z.~Chen and H.~Zhang, ``Learning implicit fields for generative shape
  modeling,'' in \emph{Proc. CVPR}, 2019, pp. 5939--5948.

\bibitem{dong2015image}
C.~Dong, C.~C. Loy, K.~He, and X.~Tang, ``Image super-resolution using deep
  convolutional networks,'' \emph{IEEE Trans. Pattern Anal. Mach. Intell.},
  vol.~38, no.~2, pp. 295--307, 2015.

\bibitem{kim2016accurate}
J.~Kim, J.~K. Lee, and K.~M. Lee, ``Accurate image super-resolution using very
  deep convolutional networks,'' in \emph{Proc. CVPR}, 2016, pp. 1646--1654.

\bibitem{yu2018pu}
L.~Yu, X.~Li, C.-W. Fu, D.~Cohen-Or, and P.-A. Heng, ``Pu-net: Point cloud
  upsampling network,'' in \emph{Proc. CVPR}, 2018, pp. 2790--2799.

\bibitem{li2019pu}
R.~Li, X.~Li, C.-W. Fu, D.~Cohen-Or, and P.-A. Heng, ``Pu-gan: a point cloud
  upsampling adversarial network,'' in \emph{Proc. ICCV}, 2019, pp. 7203--7212.

\bibitem{qian2020pugeo}
Y.~Qian, J.~Hou, S.~Kwong, and Y.~He, ``Pugeo-net: A geometry-centric network
  for 3d point cloud upsampling,'' in \emph{Proc. ECCV}, 2020, pp. 752--769.

\bibitem{bertsekas1990auction}
D.~P. Bertsekas, ``The auction algorithm for assignment and other network flow
  problems: A tutorial,'' \emph{Interfaces}, vol.~20, no.~4, pp. 133--149,
  1990.

\bibitem{hu2021learning}
Q.~Hu, B.~Yang, L.~Xie, S.~Rosa, Y.~Guo, Z.~Wang, N.~Trigoni, and A.~Markham,
  ``Learning semantic segmentation of large-scale point clouds with random
  sampling,'' \emph{IEEE Trans. Pattern Anal. Mach. Intell.}, 2021.

\bibitem{chang2015shapenet}
A.~X. Chang, T.~Funkhouser, L.~Guibas, P.~Hanrahan, Q.~Huang, Z.~Li,
  S.~Savarese, M.~Savva, S.~Song, H.~Su \emph{et~al.}, ``Shapenet: An
  information-rich 3d model repository,'' \emph{arXiv preprint
  arXiv:1512.03012}, 2015.

\bibitem{uy2019revisiting}
M.~A. Uy, Q.-H. Pham, B.-S. Hua, T.~Nguyen, and S.-K. Yeung, ``Revisiting point
  cloud classification: A new benchmark dataset and classification model on
  real-world data,'' in \emph{Proc. ICCV}, 2019, pp. 1588--1597.

\bibitem{armeni20163d}
I.~Armeni, O.~Sener, A.~R. Zamir, H.~Jiang, I.~Brilakis, M.~Fischer, and
  S.~Savarese, ``3d semantic parsing of large-scale indoor spaces,'' in
  \emph{Proc. CVPR}, 2016, pp. 1534--1543.

\bibitem{landrieu2018large}
L.~Landrieu and M.~Simonovsky, ``Large-scale point cloud semantic segmentation
  with superpoint graphs,'' in \emph{Proc. CVPR}, 2018, pp. 4558--4567.

\bibitem{jiang2019hierarchical}
L.~Jiang, H.~Zhao, S.~Liu, X.~Shen, C.-W. Fu, and J.~Jia, ``Hierarchical
  point-edge interaction network for point cloud semantic segmentation,'' in
  \emph{Proc. ICCV}, 2019, pp. 10\,433--10\,441.

\bibitem{wang2020deep}
Z.~Wang, J.~Chen, and S.~C. Hoi, ``Deep learning for image super-resolution: A
  survey,'' \emph{IEEE Trans. Pattern Anal. Mach. Intell.}, vol.~43, no.~10,
  pp. 3365--3387, 2020.

\bibitem{yifan2019patch}
W.~Yifan, S.~Wu, H.~Huang, D.~Cohen-Or, and O.~Sorkine-Hornung, ``Patch-based
  progressive 3d point set upsampling,'' in \emph{Proc. CVPR}, 2019, pp.
  5958--5967.

\bibitem{DBLP:journals/tog/BommesZK09}
D.~Bommes, H.~Zimmer, and L.~Kobbelt, ``Mixed-integer quadrangulation,''
  \emph{ACM Trans. Graph.}, vol.~28, no.~3, p.~77, 2009.

\bibitem{DBLP:journals/cgf/LiuZXGG08}
L.~Liu, L.~Zhang, Y.~Xu, C.~Gotsman, and S.~J. Gortler, ``A local/global
  approach to mesh parameterization,'' \emph{Comput. Graph. Forum}, vol.~27,
  no.~5, pp. 1495--1504, 2008.

\bibitem{DBLP:journals/tog/SpringbornSP08}
B.~Springborn, P.~Schr{\"{o}}der, and U.~Pinkall, ``Conformal equivalence of
  triangle meshes,'' \emph{ACM Trans. Graph.}, vol.~27, no.~3, p.~77, 2008.

\bibitem{DBLP:journals/tog/ShefferLMB05}
A.~Sheffer, B.~L{\'{e}}vy, M.~Mogilnitsky, and A.~Bogomyakov, ``{ABF++:} fast
  and robust angle based flattening,'' \emph{ACM Trans. Graph.}, vol.~24,
  no.~2, pp. 311--330, 2005.

\bibitem{DBLP:journals/cgf/ZhaoSLZYLWGG20}
H.~Zhao, K.~Su, C.~Li, B.~Zhang, L.~Yang, N.~Lei, X.~Wang, S.~J. Gortler, and
  X.~Gu, ``Mesh parametrization driven by unit normal flow,'' \emph{Comput.
  Graph. Forum}, vol.~39, no.~1, pp. 34--49, 2020.

\bibitem{DBLP:journals/tvcg/JinKLG08}
M.~Jin, J.~Kim, F.~Luo, and X.~Gu, ``Discrete surface ricci flow,'' \emph{IEEE
  Trans. Vis. Comput. Graph.}, vol.~14, no.~5, pp. 1030--1043, 2008.

\bibitem{schwarz2018emerging}
S.~Schwarz, M.~Preda, V.~Baroncini, M.~Budagavi, P.~Cesar, P.~A. Chou, R.~A.
  Cohen, M.~Krivoku{\'c}a, S.~Lasserre, Z.~Li \emph{et~al.}, ``Emerging mpeg
  standards for point cloud compression,'' \emph{IEEE J. Emerg. Sel. Topic
  Circuits Syst.}, vol.~9, no.~1, pp. 133--148, 2018.

\bibitem{hou2014highly}
J.~Hou, L.-P. Chau, M.~Zhang, N.~Magnenat-Thalmann, and Y.~He, ``A highly
  efficient compression framework for time-varying 3-d facial expressions,''
  \emph{IEEE Trans. Circuits Syst. Video Technol.}, vol.~24, no.~9, pp.
  1541--1553, 2014.

\bibitem{hou2014compressing}
J.~Hou, L.-P. Chau, N.~Magnenat-Thalmann, and Y.~He, ``Compressing 3-d human
  motions via keyframe-based geometry videos,'' \emph{IEEE Trans. Circuits
  Syst. Video Technol.}, vol.~25, no.~1, pp. 51--62, 2015.

\end{thebibliography}


\end{document}